\documentclass[letterpaper]{article} 
\usepackage{aaai25}  
\usepackage{times}  
\usepackage{helvet}  
\usepackage{courier}  
\usepackage[hyphens]{url}  
\usepackage{graphicx} 
\usepackage{natbib}  
\usepackage{caption} 
\usepackage{algorithm}
\usepackage{algorithmic}

\usepackage{newfloat}
\usepackage{listings}
\DeclareCaptionStyle{ruled}{labelfont=normalfont,labelsep=colon,strut=off} 
\floatstyle{ruled}
\newfloat{listing}{tb}{lst}{}
\floatname{listing}{Listing}
\usepackage{amsfonts}
\usepackage{amsmath}
\newtheorem{theorem}{Theorem}
\newtheorem{lemma}[theorem]{Lemma}

\title{Predictive Modeling of Homeless Service Assignment: \\ A Representation Learning Approach}
\author{
   Khandker Sadia Rahman,
   Charalampos Chelmis  
}
\affiliations{
    Department of Computer Science, University at Albany, SUNY \\
    Albany, New York, USA\\
    \{srahman2, cchelmis\}@albany.edu}

\begin{document}

\maketitle

\begin{abstract}
In recent years, there has been growing interest in leveraging machine learning for homeless service assignment. However, the categorical nature of administrative data recorded for homeless individuals hinders the development of accurate machine learning methods for this task. This work asserts that deriving latent representations of such features, while at the same time leveraging underlying relationships between instances is crucial in algorithmically enhancing the existing assignment decision--making process. Our proposed approach learns temporal and functional relationships between services from historical data, as well as unobserved but relevant relationships between individuals to generate features that significantly improve the prediction of the next service assignment compared to the state--of--the--art.
\end{abstract}

%
\begin{links}
    \link{Code}{https://github.com/IDIASLab/REPLETE}
\end{links}

\section{Introduction}

Machine learning has gained significant attention for its ability to solve complex real--world problems across various domains including criminal justice, e--commerce, healthcare, banking, finance, and social service \cite{sarker2021machine}. At the same time, a constant rise in the rate of homelessness has been observed \cite{dej2020turning}. In the United States alone, homelessness has risen $12.5\%$ between $2022$ and $2023$, with a particularly sharp increase of $29.5\%$ in chronic patterns of homelessness \cite{ahar2023}. Chronic pattern of homelessness occurs when individuals experience repeated homelessness over an extended period (e.g., at least two years), or endure continuous homelessness for at least 12 months \cite{fleury2021met}.

To date, various machine learning approaches have been proposed for predicting whether individuals will reenter the homeless system \cite{gao2017homelessness, hong2018applications, kube2019allocating} as well as assessing the risk of chronic homelessness of individuals \cite{vanberlo2021interpretable, messier2021predicting}. However, only limited work has addressed the more challenging problem of homeless service assignment \cite{rahman2022learning,rahman2023bayesian, pokharel2024discretionary}. Specifically, \cite{pokharel2024discretionary} investigate simple  models (e.g., Decision Trees with Short Explainable Rules) to recommend the exact service individuals can benefit from, while \cite{rahman2022learning, rahman2023bayesian} explicitly model the trajectories of individuals within the homeless system. The main limitation of these methods lies in their inability to capture the informative relationships between different feature values (i.e., within--feature interactions) and the complex between--feature interactions.

Administrative data collected by the homeless service providers comprises of $(i)$ service assignments and duration of stay, $(ii)$ housing outcome after service assignments, $(iii)$ demographics (e.g., race, gender, ethnicity), educational history, disabling condition, and $(iv)$ time--variant information (e.g., monthly income, health). Most of these features are categorical, and existing methods use one--hot encoding to handle them, mainly for simplicity. Unfortunately, such treatment misses the rich relationships within the features, while exponentially increasing the dimensionality of the data with growing number of categories \cite{rodriguez2018beyond}. Instead, we propose a predictive model that learns latent representations for features and services, which are then utilized to capture feature interactions and relationships between individuals. We show that the learned representations significantly improve the accuracy of homeless service assignment compared to the state--of--the--art. 

This brings us to the core contribution of our work. Similar to \cite{rahman2022learning,rahman2023bayesian} we study the interactions between services found in individuals' history. However, those studies oversimplify the task of predicting the likelihood of reaching the next service based solely on transitional probabilities between past services, while we assert that many other interactions exist that provide valuable context which can in turn enhance predictive performance. Specifically, subsequent services are often assigned to individuals to serve a particular purpose at a specific time. Understanding such functional and temporal relationship between services can provide insights into why a particular service is assigned at a specific time. Our work extends beyond previous studies by incorporating temporal and functional interaction between services. Furthermore, our model explores the relationships between individuals. Generally, individuals with similar features tend to receive similar services, and leveraging such information can significantly improve the robustness of predictive models. While \cite{rahman2022learning} incorporates finding the ``most similar individual'' in their prediction problem, it does so as an information retrieval task; instead our approach clusters individuals and utilizes the collective patterns within clusters to provide better insights and more accurate predictions.

In summary, given the history of services assigned to individuals and their current socio--economic features, we propose an objective function with three distinct components that captures $(i)$ temporal and $(ii)$ functional interactions between services, and $(iii)$ interactions between individuals (instances). Through optimization, we obtain latent representations for services and features, and identify the clusters to which each individual belongs. Next, features derived from these representations are input into a feed--forward neural network for predicting the next service assignment. We state our main contributions as follows:

\begin{enumerate}
    \item We propose a representation--based predictive model for homeless service assignment that effectively learns the latent representation of services and features.
    
    \item We design a novel optimization function that incorporates temporal, functional, and instance interactions to enhance the learning of representations.

    \item We utilize the learnt representations to derive features that significantly improve the performance of service assignment predictions.

\end{enumerate}

The rest of the paper is organized as follows. Section \ref{sec:related} summarizes the related work. Section \ref{sec:problem} delineates the problem setting. Section \ref{sec:replete} introduces the proposed approach. Section \ref{sec:exp} describes the data, metrics, and baselines. Section \ref{sec:results} provides detailed discussion of the experimental results. Section \ref{sec:conclude} concludes with a discussion of the limitations of our study and potential directions for future work.

\section{Related Work}
\label{sec:related}
\paragraph{Reentry Prediction and Risk of Chronic Homelessness\\}
A considerable body of prior work has focused on predicting reentry into homelessness and assessing the risk of homelessness or chronic homelessness. \cite{gao2017homelessness,hong2018applications} investigated the application of machine learning models such as logistic regression and random forest to predict whether individuals will reenter homelessness. On the other hand, \cite{vanberlo2021interpretable,messier2021predicting} explored neural networks to assess the risk of chronic homelessness. Additionally, \cite{vajiac2024preventing} investigated various machine learning models to evaluate the risk of eviction--caused homelessness and accurately identify individuals in need of assistance. All of these works are formulated as binary classification problems. Our work, on the other hand, focuses on developing a predictive model for service assignment task, which is a more challenging multi--class classification problem. 

\paragraph{Homeless service assignment\\}
Another body of work focuses on predicting the service assignment at various conditions, such as \cite{toros2018prioritizing, shinn2013efficient, greer2016targeting} develop models that prioritize housing for those at high risk of homelessness. In contrast, our work centers on individuals who have already experienced chronic patterns of  homelessness. On the other hand, \cite{chelmis2021smart} explores machine learning models to predict the exact support corresponding to the assigned service. Similary, \cite{pokharel2024discretionary} employs decision trees with simple explainable rules (SER--DT) algorithm, which divides the problem into multiple binary classification tasks using a one--versus--all prediction approach for service assignment upon entry. While these studies use one--hot encoding for categorical features, in contrast, our approach leverages representation learning to uncover the rich relationships within these features, moving beyond one--hot encoding. Our research aligns with \cite{rahman2022learning} and \cite{rahman2023bayesian}  which address the multi--class service assignment task.  Specifically, \cite{rahman2022learning} infers a homelessness network and \cite{rahman2023bayesian} utilizes a Bayesian network to predict the next service assignment. In contrast, our work integrates both history of individuals within the homeless system and socio--economic features to predict the next service assignment. A key distinction is that while previous studies explore simple transitional relationships between services, our approach leverages more complex (temporal and functional) relationships between services, and relevant interactions between individuals to predict the next service assignment.

\section{Problem Setting}
\label{sec:problem}
We denote the set of individuals with chronic patterns of homelessness as $\mathcal{U} = \{ u_1, \dots, u_{|\mathcal{U}|}\}$. The set of homeless services, such as permanent housing, rapid rehousing, day shelter, is denoted by $\mathcal{A}=\{a^1,a^2,\dots,a^{|\mathcal{A}|}\}$. For each $u_i \in \mathcal{U}$, let $\mathcal{T}_{u_i} = \{ (a_{t_1}, t_1), \dots, (a_{t_N}, t_N)\}$ represent the history of services assigned to individual $u_i$, where $a_{t_i}$ signifies assignment to a service at time $t_i$. Next, we denote $\mathbf{X} \in \mathbb{R}^{|\mathcal{U}|\times |\mathcal{F}|}$ as the feature matrix which comprises the feature set $\mathcal{F} \in \mathbb{R}^{d}$ of individuals $u_i\in \mathcal{U}$ at time $t_{N+1}$. 

Given matrix $\mathbf{X}$
 and the history of prior service assignments $\mathcal{T}_{u_i} = \{ (a_{t_1}, t_1), \dots, (a_{t_N}, t_N)\}$ for individuals $u_i\in\mathcal{U}$, the goal of this paper is to accurately predict the next service assignment $a_{t_{N+1}} \in \mathbf{Y}$ for $\forall u_i$.

\section{Proposed Approach}
\label{sec:replete}
\begin{figure}[t!]
    \centerline{\includegraphics[width=\columnwidth]{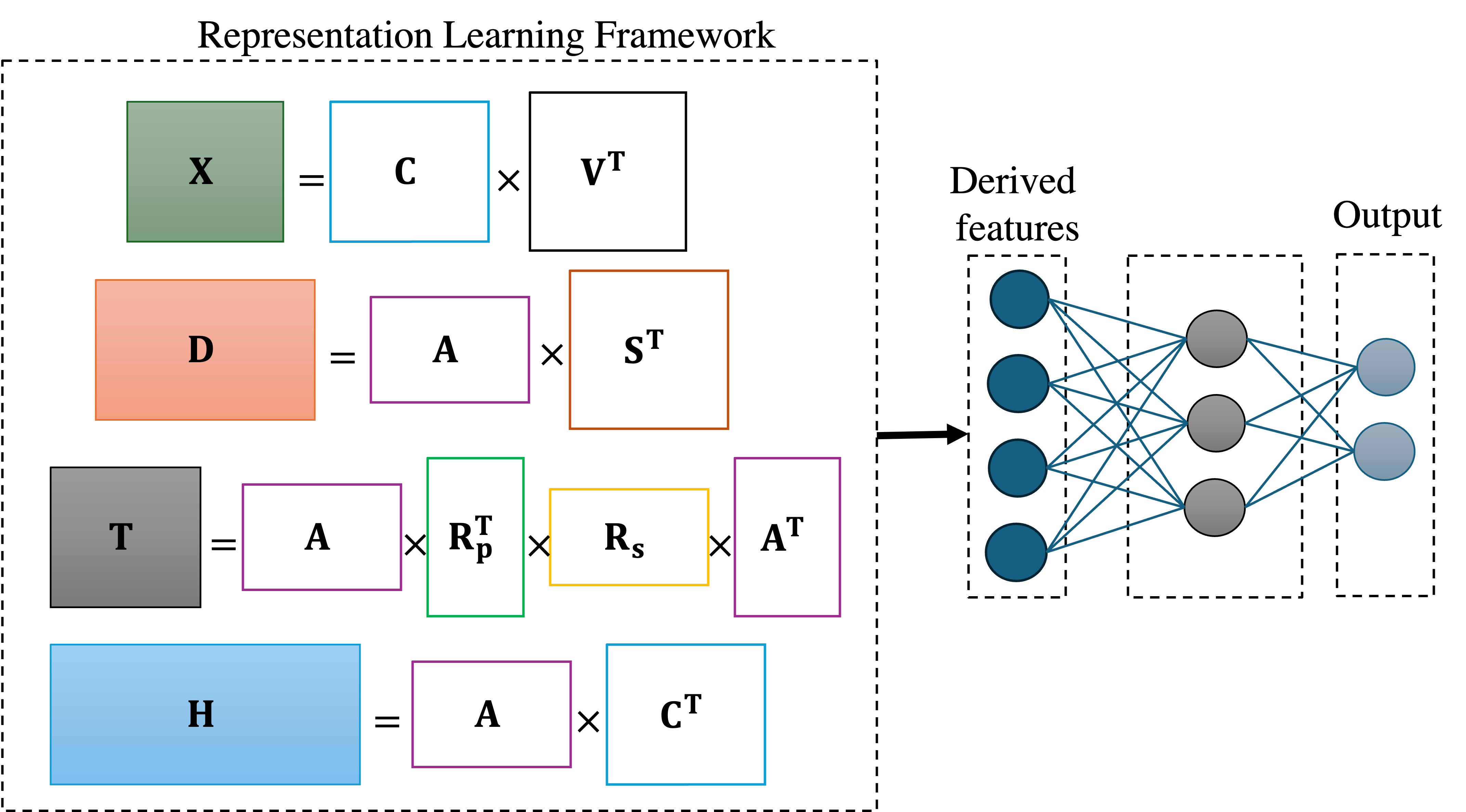}}
    \caption{Overview of REPLETE. Representation learning framework first learns representations $\mathbf{A,C,V,S,R_p,R_s}$. These are used to derive features, which are subsequently input into FFNN for service assignment prediction.} 
    \label{fig:overview}
\end{figure}

In this section, we present a comprehensive overview of REPLETE, a \underline{REP}resentation \underline{L}earning--based s\underline{E}rvice assignmen\underline{T} mod\underline{E}l, which consists of two main components: $(i)$ representation learning framework and $(ii)$ prediction model. The representation learning framework is designed to learn the latent representations of services. The prediction model leverages the learned representations to enhance the homeless service assignment decision--making process. Figure \ref{fig:overview} provides an overview of REPLETE.

\subsection{Representation Learning Framework}
\label{sec:learningframework}
To effectively learn the representations of services, we design an optimization function that captures three different contexts. The first context focuses on the \emph{temporal} interactions between services, where assignments at different stages in the homeless system are likely to have distinct vector representations. The second context captures the \emph{functional} interactions between services, where assignments serving similar needs share similar vector representations. The last context models relationships between \emph{individuals}, where similar individuals are more likely to exhibit similar traits.

\paragraph{Temporal Context}
Here, our goal is to learn the latent representations of services and time units based on the intuitive observation that services occurring within similar time units should have analogous vector representations. To formalize this, we define $\mathbf{D} \in \mathbb{R}^{|\mathcal{A}| \times |\tau|}$, where $\mathbf{D}_{ij}$ represents the frequency with which service $a^i$ appears in time unit $\tau_j$. Here, $\tau$ denotes the set of time units (e.g., weeks, months, or years). To derive standardized representations, we employ Non--negative Matrix Factorization (NMF). Specifically, given $\mathbf{D}$, we learn the non--negative matrices, $\mathbf{A} \in \mathbb{R}^{|\mathcal{A}|\times k}$ for services and $\mathbf{S}\in\mathbb{R}^{|\tau|\times k}$ for time units by solving
 \begin{equation*}
        \begin{split}
        \min_{\mathbf{A}, \mathbf{S} \geq 0}\ & \| \mathbf{D} - \mathbf{A}\mathbf{S}^{\top} \|_F^2 +  \alpha \sum_{j=1}^{|\tau|-1}{\| \mathbf{S}_{j,:} -\mathbf{S}_{j-1,:} \|_F^2} \\ 
        & + \lambda (\|\mathbf{A}\|_F^2 + \|\mathbf{S}\|_F^2),
        \end{split}
        \label{eq:temporal}
    \end{equation*}
where $\alpha$, $\lambda$, and $k$ control the smoothness, sparsity constraints, and dimension of representations, respectively.

\paragraph{Functional Context} This context is based on the premise that subsequent services are assigned to individuals because they address needs unmet by the preceding service, meaning, the two services together complement each other in satisfying a need. In this context, we capture the latent relationships between such service pairs. We define a transitional probability matrix $\mathbf{T}\in \mathbb{R}^{|\mathcal{A}| \times |\mathcal{A}| }$, where each element $\mathbf{T}_{ij}$ denotes the probability of assigning service $a^j$ after service $a^i$ \cite{rahman2022learning}. Consequently, we learn $m$ latent relations between services represented by matrices, $\mathbf{R_p} \in \mathbb{R}^{m \times k}$ for preceding assignment and $\mathbf{R_s} \in \mathbb{R}^{m \times k}$ for succeeding assignments such that for any pair $(a^i, a^j)$, $\mathbf{T}_{ij}=\sum_{l=1}^m{a^i \mathbf{R}_p[l] + a^j \mathbf{R}_s[l]}$. We perform symmetric NMF \cite{kuang2012symmetric} to obtain the solution to
\begin{equation*}
        \begin{split}
        \min_{\mathbf{A, R_p, R_s} \geq 0} &  \| \mathbf{T} - \mathbf{A} \mathbf{R_p}^{\top} \mathbf{R_s} \mathbf{A}^{\top} \|_F^2 + \\ 
         & + \lambda (\|\mathbf{A}\|_F^2 + \|\mathbf{R_p}\|_F^2 + \|\mathbf{R_s}\|_F^2),
        \end{split}
        \label{eq:functional}
    \end{equation*}
where $\mathbf{A}$ denotes the latent representations of services. Similar to the temporal context, we incorporate a sparsity constraint with $\lambda$ controlling the penalty for overfitting, and $k$ denoting the dimension of the representations.

\paragraph{Individual Context}
In this context, our objective is to learn the latent representations of services denoted by $\mathbf{A}$ and features denoted by $\mathbf{V} \in \mathbb{R}^{|\mathcal{F}|\times k}$, while also identifying clusters to which individuals belong (represented by $\mathbf{C} \in \mathbb{R}^{|\mathcal{U}|\times k}$). By clustering individuals with similar features and past service assignments, our goal is to ensure that services assigned to similar individuals (i.e., in the same cluster) have similar representations. We define $\mathbf{H} \in \mathbb{R}^{|\mathcal{A}| \times |\mathcal{U}|}$, where each element $\mathbf{H}_{ij}$ denotes the number of times individual $u_j$ is assigned to service $a^i$. Since $\mathbf{X}$ and $\mathbf{H}$ exhibit higher sparsity, $L_{2-1}$ norm is applied to prevent rows with greater sparsity from dominating the objective function. Given matrices $\mathbf{H}$ and $\mathbf{X}$, we solve 
\begin{equation*}
    \begin{split}
       \min_{\mathbf{A}, \mathbf{V}\geq 0, \mathbf{C}}
        &\| \mathbf{H} - \mathbf{A}\mathbf{C}^{\top} \|_{2,1} + \|\mathbf{X} - \mathbf{C}\mathbf{V}^{\top} \|_{2,1}+\lambda (\|\mathbf{A}\|_F^2\\
       & + \|\mathbf{V}\|_F^2+\|\mathbf{C}\|_F^2 ) +\beta tr(\mathbf{C}^{\top}\mathbf{\Gamma}\mathbf{C}) \\
       & \textit{s.t.  } \mathbf{C}\mathbf{1}_k=\mathbf{1}_{|\mathcal{U}|}, \mathbf{C}\in \mathbb[0,1].
    \end{split}
    \label{eq:individual}
    \end{equation*}
Here, the term $tr(\mathbf{C}^{\top}\mathbf{\Gamma}\mathbf{C})$ ensures that similar individuals are clustered together and  $\mathbf{\Gamma} \in \mathbb{R}^{|\mathcal{U}|\times |\mathcal{U}|}$ records the cosine similarity between individuals in the set $\mathcal{U}$. The constraints on $\mathbf{C}$ enforce soft clustering by ensuring that the rules of probability are satisfied. Such soft clustering is appropriate for homelessness, a complex domain where individuals can exhibit similarities with members of different clusters. Finally, $\beta$ controls the strictness of the clusters by determining how tightly or loosely data instances are grouped together.

\paragraph{Overall Optimization Function}
With all the previously introduced components, we formulate our joint optimization problem as follows:
\begin{equation}
        \begin{split}
        &\min_{\mathbf{A, S, V, R_p, R_s} \geq 0, C}   \| \mathbf{D} - \mathbf{AS}^{\top} \|_F^2 + \alpha \sum_{j=1}^{|\tau|-1}{\| \mathbf{S}_{j,:} -\mathbf{S}_{j-1,:} \|_F^2} \\
        & + \| \mathbf{H} - \mathbf{A}\mathbf{C}^{\top} \|_{2,1} + \|\mathbf{X} - \mathbf{C}\mathbf{V}^{\top} \|_{2,1}+\beta tr(\mathbf{C}^{\top}\mathbf{\Gamma}\mathbf{C}) \\
       &+  \| \mathbf{T} - \mathbf{A R_p}^{\top}\mathbf{ R_s A}^{\top} \|_F^2+ \lambda (\|\mathbf{A}\|_F^2 + \|\mathbf{S}\|_F^2 + \|\mathbf{V}\|_F^2  \\
        &+\|\mathbf{C}\|_F^2+ \|\mathbf{R_p}\|_F^2 + \|\mathbf{R_s}\|_F^2)\\
        & \textit{s.t.  } \mathbf{C}\mathbf{1}_k=\mathbf{1}_{|\mathcal{U}|}, \mathbf{C}\in \mathbb[0,1].
        \end{split}
        \label{eq:joint}
    \end{equation}

\subsection{Optimization Algorithm}
\label{sec:opt-sol}
Jointly updating the variables in Eq.~(\ref{eq:joint}) causes the objective function to become non--convex. We therefore optimize the objective function using Alternating Direction Method of Multiplier (ADMM) \cite{boyd2011distributed}, where variables are updated separately. All proofs are provided in the Appendix section.

We begin by relaxing the constraints on $\mathbf{C}$ to orthogonality, specifically $\mathbf{C}^\top\mathbf{C}=\mathbf{I}, \mathbf{C}\ge0$ \cite{tang2012unsupervised}. We then introduce two auxiliary variables: $\mathbf{P}=\mathbf{H}-\mathbf{AC}^\top$ and $\mathbf{Q}=\mathbf{X}-\mathbf{CV}^\top$. Consequently, Eq.~(\ref{eq:joint}) is reformulated into the following equivalent problem:
\begin{equation}
        \begin{split}
        &\min_{\mathbf{A, S, V, R_p, R_s,C} \geq 0,\mathbf{P},\mathbf{Q}} \| \mathbf{D} - \mathbf{AS}^{\top} \|_F^2 + \alpha \|\mathbf{S}\mathbf{B}^{'}\|_F^2 + \| \mathbf{P}\|_{2,1} \\
        &  + \|\mathbf{Q} \|_{2,1}+\beta tr(\mathbf{C}^{\top}\mathbf{\Gamma}\mathbf{C}) +  \| \mathbf{T} - \mathbf{A R_p}^{\top}\mathbf{ R_s A}^{\top} \|_F^2 \\
        & + \lambda (\|\mathbf{A}\|_F^2 + \|\mathbf{S}\|_F^2 + \|\mathbf{V}\|_F^2 +\|\mathbf{C}\|_F^2+ \|\mathbf{R_p}\|_F^2 + \|\mathbf{R_s}\|_F^2)\\ 
        &+ \langle \mathcal{L},\mathbf{X}-\mathbf{CV}^\top-\mathbf{Q} \rangle+ \langle \mathcal{K},\mathbf{H}-\mathbf{AC}^\top-\mathbf{P} \rangle\\
        &+ \langle \mathcal{N},\mathbf{C^\top C-I} \rangle+ \frac{\mu}{2}\|\mathbf{H}-\mathbf{AC}^\top-\mathbf{P} \|_F^2\\
        &+ \frac{\mu}{2}\|\mathbf{X}-\mathbf{CV}^\top-\mathbf{Q} \|_F^2,
        \end{split}
        \label{eq:admm}
    \end{equation}
where $\mathcal{K,L,}$ and $\mathcal{N}$ are Lagrangian multipliers, $\mu$ controls the penalty for violating the equality constraints, $\langle \cdot, \cdot \rangle$ denotes the dot product, and $\mathbf{B}^{'}$ represents the matrix equivalent to the smoothing constraint. Next, we summarize the update rules for each variable. 

\paragraph{Update $\mathbf{P}$}
For updating $\mathbf{P}$, we hold the other variables constant and remove the terms that are irrelevant to $\mathbf{P}$. Eq.~(\ref{eq:admm}) can be written as:
\begin{equation}
        \min_{\mathbf{P}} \frac{1}{2}\|\mathbf{P}-(\mathbf{H}-\mathbf{A}\mathbf{C}^T+\frac{1}{\mu}\mathcal{K})\|_F^2+\frac{1}{\mu}\|\mathbf{P}\|_{2,1}   
        \label{eq:P}
\end{equation}
Lemma \ref{lemma:2,1} provides a closed form solution for Eq.~(\ref{eq:P}).
\begin{lemma} \cite{wang2015embedded}
    Given matrix $\mathbf{E}$ and a positive scale $\alpha$, 
    the $i^{th}$ row of the optimal solution $\mathbf{W}^*$ of $\min_\mathbf{W} \frac{1}{2}\|\mathbf{W}-\mathbf{E}\|_F^2+\alpha\|\mathbf{W}\|_{2,1}$ is given by:
    \begin{equation*}
        w_i^* = \left\{\begin{array}{lr} (1-\frac{\alpha}{\|e_i\|})e_i, & \|e_i\|>\alpha\\
        0, & \text{otherwise}\end{array}\right.
    \end{equation*}
    \label{lemma:2,1}
\end{lemma}
Using Lemma~\ref{lemma:2,1}, the optimal solution $\mathbf{P}^*$ for the above equation is as follows, where $\mathbf{E}^\mathbf{P}=\mathbf{H}-\mathbf{A}\mathbf{C}^T+\frac{1}{\mu}\mathcal{K}$:
\begin{equation*}
        \mathbf{P}_{i,:}^* = \left\{\begin{array}{lr} (1-\frac{1}{\mu\|\mathbf{E}^\mathbf{P}_{i,:}\|})\mathbf{E}^\mathbf{P}_{i,:}, & \|\mathbf{E}^\mathbf{P}_{i,:}\|>\frac{1}{\mu}\\
        0, & \text{otherwise}\end{array}\right.
\end{equation*}

\paragraph{Update $\mathbf{Q}$} 
The optimal solution $\mathbf{Q}^*$ is obtained similar to $\mathbf{P}$, using Lemma~\ref{lemma:2,1}, where $\mathbf{E}^\mathbf{Q}=\mathbf{X}-\mathbf{C}\mathbf{V}^T+\frac{1}{\mu}\mathcal{L}$:
\begin{equation*}
        \mathbf{Q}_{i,:}^* = \left\{\begin{array}{lr} (1-\frac{1}{\mu\|\mathbf{E}^\mathbf{Q}_{i,:}\|})\mathbf{E}^\mathbf{Q}_{i,:}, & \|\mathbf{E}^\mathbf{Q}_{i,:}\|>\frac{1}{\mu}\\
        0, & \text{otherwise}\end{array}\right.
\end{equation*}

\paragraph{Update $\mathbf{S}$}
Let $\psi_\mathbf{S}$ be the Lagrangian multiplier for $\mathbf{S}\ge0$, the Lagrangian function related to $\mathbf{S}$ is, $\mathcal{O}_\mathbf{S}=\min \|\mathbf{D}-\mathbf{A}\mathbf{S}^T\|_F^2 + \lambda \|\mathbf{S}\|_F^2 + \alpha \|\mathbf{SB'}\|_F^2 - tr(\mathbf{\psi_SS^\top})$. The partial derivative of $\mathcal{O}_\mathbf{S}$ is
    $\frac{1}{2}\frac{d\mathcal{O}_\mathbf{S}}{d\mathbf{S}}= -(\mathbf{D}-\mathbf{AS}^\top)^\top \mathbf{A}+\alpha(\mathbf{S}\mathbf{B}^{'})\mathbf{B}^{'}+\lambda\mathbf{S}-\psi_\mathbf{S}$. 
Using the Karush--KuhnTucker (KKT) complementary condition \cite{boyd2004convex}, i.e., $\psi_\mathbf{S}(i,j)\mathbf{S}_{ij}=0$, we get $\mathbf{S}_{ij} \leftarrow \mathbf{S}_{ij} \frac{\hat{\mathbf{S}}_{ij}}{\Tilde{\mathbf{S}}_{ij} }$ \cite{zhang2017unified}, where
\begin{equation*}
\begin{split}
    &\hat{\mathbf{S}}_{ij} = \mathbf{D}^{\top}\mathbf{A}+[\alpha(\mathbf{S}{(\mathbf{B}^{'})}^2)]^{-}\\
    &\Tilde{\mathbf{S}}_{ij} = \mathbf{SA}^\top\mathbf{A}+\lambda\mathbf{S}+[\alpha(\mathbf{S}{(\mathbf{B}^{'})}^2)]^{+}.
\end{split}
\end{equation*}
Here, for any matrix $\mathbf{M}$, $(\mathbf{M})^{+} = \frac{ABS(\mathbf{M})+\mathbf{M}}{2}$ and $(\mathbf{M})^{-} =\frac{ABS(\mathbf{M})-\mathbf{M}}{2}$ are the positive and negative part of $\mathbf{M}$, respectively, and $ABS(\mathbf{M})$ consists of the absolute value of elements in $\mathbf{M}$ \cite{shu2019beyond}.

\paragraph{Update $\mathbf{R}_p$ and $\mathbf{R}_s$} The partial derivative of the Lagrangian objective function w.r.t. $\mathbf{R}_p$ is
$    \frac{1}{2}\frac{d\mathcal{O}_\mathbf{R_p}}{d\mathbf{R_p}}= -\mathbf{T}\mathbf{A}\mathbf{A}^\top\mathbf{R_s} + \mathbf{A R_p}^{\top}\mathbf{ R_s A}^{\top}\mathbf{A}\mathbf{A}^\top\mathbf{R_s}
    +\lambda \mathbf{R_p}-\psi_\mathbf{R_p}$. 
Using the KKT complementary condition, we get $\mathbf{R}_{\mathbf{p}_{ij}} \leftarrow \mathbf{R}_{\mathbf{p}_{ij}} \frac{\hat{\mathbf{R}}_{\mathbf{p}_{ij}}}{\Tilde{\mathbf{R}}_{\mathbf{p}_{ij}}}$, where
\begin{equation*}
\begin{split}
    &\hat{\mathbf{R}}_{\mathbf{p}_{ij}} = \mathbf{T}\mathbf{A}\mathbf{A}^\top\mathbf{R_s}+(\mathbf{A R_p}^{\top}\mathbf{ R_s A}^{\top}\mathbf{A}\mathbf{A}^\top\mathbf{R_s})^{-}\\
    &\Tilde{\mathbf{R}}_{\mathbf{p}_{ij}} = \lambda \mathbf{R_p}+(\mathbf{A R_p}^{\top}\mathbf{ R_s A}^{\top}\mathbf{A}\mathbf{A}^\top\mathbf{R_s})^{+}.
\end{split}
\end{equation*}
Similar to $\mathbf{R_p}$, we get $\mathbf{R}_{\mathbf{s}_{ij}} \leftarrow \mathbf{R}_{\mathbf{s}_{ij}} \frac{\hat{\mathbf{R}}_{\mathbf{s}_{ij}}}{\Tilde{\mathbf{R}}_{\mathbf{s}_{ij}}}$ for $\mathbf{R_s}$, such that:
\begin{equation*}
\begin{split}
    &\hat{\mathbf{R}}_{\mathbf{s}_{ij}} = \mathbf{T}\mathbf{A}\mathbf{A}^\top\mathbf{R_p}+(\mathbf{A R_p}^{\top}\mathbf{ R_s A}^{\top}\mathbf{A}\mathbf{A}^\top\mathbf{R_p})^{-}\\
    &\Tilde{\mathbf{R}}_{\mathbf{s}_{ij}} = \lambda \mathbf{R_s}+(\mathbf{A R_p}^{\top}\mathbf{ R_s A}^{\top}\mathbf{A}\mathbf{A}^\top\mathbf{R_p})^{+}.
\end{split}
\end{equation*}
\paragraph{Update $\mathbf{V}$} The partial derivative of the Lagrangian objective function w.r.t. $\mathbf{V}$ is $\frac{d\mathcal{O}_\mathbf{V}}{d\mathbf{V}}= -\mu(\mathbf{X}-\mathbf{CV}^\top-\mathbf{Q})^\top\mathbf{C}-\mathcal{L}^\top\mathbf{C} +2\lambda\mathbf{V} -\psi_\mathbf{V}$. Using the KKT complementary condition, we get $\mathbf{V}_{ij} \leftarrow \mathbf{V}_{ij} \frac{\hat{\mathbf{V}}_{ij}}{\Tilde{\mathbf{V}}_{ij}}$, where
\begin{equation*}
\begin{split}
    &\hat{\mathbf{V}}_{ij} = \mu\mathbf{X}^\top\mathbf{C} + (\mathcal{L}^\top\mathbf{C})^{+}\\
    &\Tilde{\mathbf{V}}_{ij} = \mu\mathbf{VC}^\top\mathbf{C}+\mu\mathbf{Q}^\top\mathbf{C}+(\mathcal{L}^\top\mathbf{C})^{-} +2\lambda\mathbf{V}.
\end{split}
\end{equation*}
\paragraph{Update $\mathbf{C}$}
The partial derivative of the Lagrangian objective function w.r.t. $\mathbf{C}$ is $\frac{1}{2}\frac{d\mathcal{O}_\mathbf{C}}{d\mathbf{C}}= -\mu(\mathbf{H}-\mathbf{AC}^\top-\mathbf{P})^\top\mathbf{A}-\mu(\mathbf{X}-\mathbf{CV}^\top-\mathbf{Q})\mathbf{V} -\mathcal{L}\mathbf{V}-\mathcal{K}^\top\mathbf{A}+2\mathbf{C}\mathcal{N}+2\beta\mathbf{\Gamma}\mathbf{C}+2\lambda\mathbf{C}-\psi_\mathbf{C}$. Using the KKT complementary condition, we get $\mathbf{C}_{ij} \leftarrow \mathbf{C}_{ij} \frac{\hat{\mathbf{C}}_{ij}}{\Tilde{\mathbf{C}_{ij}}}$, where
\begin{equation*}
\begin{split}
    &\hat{\mathbf{C}}_{ij} = \mu\mathbf{H}^\top\mathbf{A}+\mu\mathbf{X}\mathbf{V}+(\mathcal{L}\mathbf{V})^{+}+(\mathcal{K}^\top\mathbf{A})^{+}+2(\mathbf{C}\mathcal{N})^{-}\\
    &+2(\beta(\mathbf{\Gamma}\mathbf{C})^{-}\\
    &\Tilde{\mathbf{C}}_{ij} = \mu\mathbf{C}\mathbf{A}^\top\mathbf{A}+\mu \mathbf{P}^\top\mathbf{A}+\mu\mathbf{C}\mathbf{V}^\top\mathbf{V}+\mu\mathbf{Q}\mathbf{V}+(\mathcal{L}\mathbf{V})^{-}\\
    &+(\mathcal{K}^\top\mathbf{A})^{-} +2(\mathbf{C}\mathcal{N})^{+}+2\beta(\mathbf{\Gamma}\mathbf{C})^{+}+2\lambda\mathbf{C}.
\end{split}
\end{equation*}

\paragraph{Update $\mathbf{A}$} 
The partial derivative of the Lagrangian objective function w.r.t. $\mathbf{A}$ is $\frac{1}{2}\frac{d\mathcal{O}_\mathbf{A}}{d\mathbf{A}}= -2(\mathbf{D}-\mathbf{AS}^\top)\mathbf{S}-4(\mathbf{T}-\mathbf{A R_p}^{\top}\mathbf{ R_s A}^{\top} ) \times \mathbf{A R_p}^{\top}\mathbf{ R_s }+2\lambda\mathbf{A}-\mu(\mathbf{H}-\mathbf{A}\mathbf{C}^\top-\mathbf{P})\mathbf{C}-\mathbf{K}\mathbf{C}-\psi_\mathbf{A}$. Using the KKT complementary conditions, we get $\mathbf{A}_{ij} \leftarrow \mathbf{A}_{ij} \frac{\hat{\mathbf{A}}_{ij}}{\Tilde{\mathbf{A}_{ij}}}$ where
\begin{equation*}
\begin{split}
    &\hat{\mathbf{A}}_{ij} =2\mathbf{D}\mathbf{S}+4\textbf{T}\mathbf{A}\mathbf{R_p}^\top\mathbf{R_s}+4(\mathbf{A R_p}^{\top}\mathbf{ R_s A}^{\top}\mathbf{A R_p}^{\top}\mathbf{ R_s })^{-}\\ &+\mu\mathbf{H}\mathbf{C}+ (\mathbf{K}\mathbf{C})^{+}\\
    &\Tilde{\mathbf{A}}_{ij} = 2 \mathbf{AS}^\top\mathbf{S}+4(\mathbf{A R_p}^{\top}\mathbf{ R_s A}^{\top}\mathbf{A R_p}^{\top}\mathbf{ R_s })^{+}+2\lambda\mathbf{A}\\
    &+\mu\mathbf{A}\mathbf{C}^\top\mathbf{C}+\mu\mathbf{P}\mathbf{C}+(\mathbf{K}\mathbf{C})^{-}.
\end{split}
\end{equation*}

\paragraph{Update $\mathcal{L}$, $\mathcal{M}$, and $\mathcal{N}$}
After updating the variables, we update the Lagrangian parameters \cite{wang2015embedded} as follows:
\begin{equation*}
    \begin{split}
        &\mathcal{L} = \mathcal{L} + \mu(\mathbf{X}-\mathbf{CV}^\top-\mathbf{Q}) \\
        &\mathcal{M} = \mathcal{M} + \mu(\mathbf{H}-\mathbf{AC}^\top-\mathbf{P}) \\
        &\mathcal{N} = \mathcal{N} + \mu(\mathbf{C^\top C-I}).
    \end{split}
\end{equation*}

\subsection{Service Assignment Prediction}
In this section, we delve into the specifics of the service assignment module of REPLETE. Once the representations are obtained by solving Eq.~(\ref{eq:joint}), we derive three sets of features: $(i)$ service and feature representations, $(ii)$ feature interactions, and $(iii)$ instance interactions. The overall process is summarized in Algorithm \ref{alg:overview}.

\paragraph{Service and Feature Representations} We utilize the representations of services and features in our prediction model instead of one--hot encoding them. The rational is that by treating services and features as categorical our model would miss the inherent relationships within them. Therefore, we replace the services $\{a_{t_1},\dots,a_{t_N}\}$ within the history of services received by an individual, with their respective representations $\{\mathbf{A}_{a_{t_1}},\dots,\mathbf{A}_{a_{t_N}}\}$, where $\mathbf{A}_{a_{t_i}}$ denotes the representation of service $a_{t_i}$. We additionally obtain the representations of the socio--economic features for each individual by multiplying $\mathbf{X}$ and $\mathbf{V}$, where $\mathbf{X}$ is the feature matrix and $\mathbf{V}$ is the matrix of learned feature representations.

\paragraph{Feature Interactions} Beyond the representations of services and features, we further incorporate two categories of service interactions: temporal and functional (Section~\ref{sec:learningframework}). Given the sequence of services $\{a_{t_1},\dots,a_{t_N}\}$ received by individuals, the temporal and functional interactions within each service pair $(a_{t_i},a_{t_j})$ are defined as  $\mathbf{A}_{a_{t_i}}\mathbf{S}^\top\mathbf{S}\mathbf{A}_{a_{t_j}}^\top$ and $\mathbf{A}_{a_{t_i}}\mathbf{R}_p^\top\mathbf{R}_s\mathbf{A}_{a_{t_j}}^\top$, respectively. Matrix $\mathbf{A}_{a_{t_i}}$ denotes the representation of service $a_{t_i}$, whereas matrices $\mathbf{S}$, $\mathbf{R}_p$, and $\mathbf{R}_s$ are the learned representations for time units, preceding assignments, and succeeding assignments, respectively.

\paragraph{Instance Interactions} We define instance interaction as the similarity between individuals, which we use to identify clusters within the individuals. Since the relaxed optimization on $\mathbf{C}$ (i.e., the matrix denoting the clusters to which individuals belong) does not directly provide probabilities for soft clustering, we normalize $\mathbf{C}$ such that each row sums to $1$, thereby obtaining these probabilities. 

\paragraph{Prediction Model}
We use $\mathbf{Z} \in \mathbb{R}^{|\mathcal{U}| \times |\mathcal{F}'|}$, to denote the concatenated feature vector, i.e., service and feature representations, feature interactions, and instance interactions, where $\mathcal{U}$ is the set of individuals, and $\mathcal{F}'$ is the set of derived features. For the service assignment task, we utilize a single hidden layer feed--forward neural network (FFNN). The rationale behind a single hidden layer architecture is that it balances complexity and computational efficiency. 

Predicted service assignments, denoted by $\mathbf{\hat{Y}}$, are calculated as $\mathbf{\hat{Y}} = \sigma(\mathbf{Z}\mathbf{W}_1+\mathbf{b}_1)\mathbf{W}_2 +\mathbf{b}_2$ where, $\mathbf{W}_1$ and $\mathbf{W}_2$ are the weight matrices for the hidden and output layers, respectively,  $\mathbf{b}_1$ and $\mathbf{b}_2$ are the bias vectors for the hidden and output layers, respectively, and $\sigma(\cdot)$ is the ReLU activation function. The model is trained using cross--entropy loss.

\begin{algorithm}
    \caption{REPLETE}
    \label{alg:overview}
    \begin{algorithmic}[1]
        \REQUIRE $\mathbf{X}$, $\mathbf{D}$, $\mathbf{T}$, $\mathbf{H}$, $\mathbf{B}^{'}$, $\alpha, \beta, \mu, \lambda, \tau, k$
        \ENSURE $\hat{a}_{t_{N+1}}$
        \STATE Randomly initialize $\mathbf{A}, \mathbf{S}, \mathbf{C}, \mathbf{V}, \mathbf{R}_p, \mathbf{R}_s, \mathbf{P}, \mathbf{Q}$
        \STATE Pre--compute similarity matrix $\mathbf{\Gamma}$
        \REPEAT 
        \STATE Update $\mathbf{A}, \mathbf{S}, \mathbf{C}, \mathbf{V}, \mathbf{R}_p, \mathbf{R}_s, \mathbf{P}, \mathbf{Q}$ according to Section \ref{sec:opt-sol}
        \UNTIL convergence
        \STATE Normalize $\mathbf{C}$ s.t. $\sum_{j} C_{i,j} = 1$
        \FOR{\textbf{each} $u_i \in \mathcal{U}$}
            \STATE $\mathbf{Z}_{u_i} = \mathbf{Z}_{u_i} \cup (\mathbf{X}\mathbf{V})_{u_i} \cup\mathbf{C}_{u_i} $  
            \FOR{\textbf{each} $a_{t_k} \in \mathcal{T}_{u_i}$}
            \STATE $\mathbf{Z}_{u_i} = \mathbf{Z}_{u_i} \cup \mathbf{A}_{a_{t_k}}$ 
            \ENDFOR
             \FOR{\textbf{each} $(a_{t_m},a_{t_n}) \in \mathcal{T}_{u_i}$}
            \STATE $\mathbf{Z}_{u_i} = \mathbf{Z}_{u_i} \cup (\mathbf{A}_{a_{t_m}}\mathbf{S}^\top\mathbf{S}\mathbf{A}_{a_{t_n}}^\top)$
            \STATE $\mathbf{Z}_{u_i} = \mathbf{Z}_{u_i} \cup (\mathbf{A}_{a_{t_m}}\mathbf{R}_p^\top\mathbf{R}_s\mathbf{A}_{a_{t_n}}^\top)$
            \ENDFOR
        \ENDFOR
        \STATE Split $\mathbf{Z}$ into training $\mathbf{Z}_{train}$ and testing set $\mathbf{Z}_{test}$
        \STATE Train FFNN with $\mathbf{Z}_{train}$
        \STATE Input $\mathbf{Z}_{test}$ into FFNN
        \STATE $\hat{\mathbf{Y}}_{test}\leftarrow$ Output of FFNN
        \RETURN $\hat{\mathbf{Y}}_{test}$
    \end{algorithmic}
\end{algorithm}

\section{Experimental Setup}
\label{sec:exp}
\paragraph{Data Description}
Our analysis utilizes an anonymized dataset from CARES of NY comprising $18,817$ records from $6,011$ chronically homeless individuals in the Capital Region of New York State, covering the period from $2012$ to $2018$, including a total of $9$ different homeless services. A complete description of the services is available at \cite{2020hmisstandards}. Sequence of services longer than $N$ are sampled using a sliding window with a length of $N$. For sequences shorter than $N$,  missing values are encoded with $0$, resulting in a zero vector representation with a length $|\mathcal{A}|$. After sampling, we denote our dataset as $\mathcal{D}$ and perform a random $70-30$ split to obtain the training set $\mathcal{D}_{train}$ and testing set $\mathcal{D}_{test}$, respectively. Finally, we train and evaluate our approach using $\mathcal{D}_{train}$ and $\mathcal{D}_{test}$ respectively. 

\paragraph{Baselines}
We compare REPLETE with the state--of--the--art: TRACE \cite{rahman2022learning} and PREVISE \cite{rahman2023bayesian}. TRACE and PREVISE uses service sequences to predict the next assignment. Additionally, following \cite{rahman2023bayesian}, we compare our approach with random forest (RF), logistic regression (LR), transformer, feed forward neural network (FFNN$_1$), recurrent neural network (RNN), and long short-term memory (LSTM), which incorporate one--hot encoded features for prediction.
\paragraph{Evaluation Metrics} We evaluate our approach using accuracy, recall, precision, and $F_1$ score. 

\paragraph{Implementation Details}
All analyses were conducted in Jupyter Notebook using Python 3 on a MacBook Air with an 8--core M1 chip (4 performance and 4 efficiency cores) and 8 GB of memory, running macOS Sonoma. Representation learning framework ran $500$ iterations, and the prediction model, on an average, ran $20$ iterations in each fold of $5$--fold cross validation.

\section{Results and Analysis}
\label{sec:results}
 \paragraph{Quantitative Analysis} Table \ref{tbl:sota} shows that REPLETE significantly outperforms the baselines in predicting the next service assignment, both in terms of accuracy and $F_1$ score. Table \ref{tbl:t-test} confirms that REPLETE significantly outperforms PREVISE (state-of-the-art) across all metrics and LSTM (the best-performing baseline) in accuracy and F-score. This indicates that $(i)$ the representation learning framework effectively captures the rich relationships within the features and instances, and $(ii)$ derived features based on temporal, functional, and individual contexts, effectively carry these relationships to the FFNN. Together, these factors lead to a significant improvement in the performance of our approach. 

\begin{table}[t]
  \centering
    \caption{Performance comparison between REPLETE and the baselines. Parameter $N$ is set to be $3$.}
	\begin{tabular}{|p{1.6cm}|c|c|c|c|}
		\hline
		  Method & Accuracy & Recall & Precision & $F_1$ score\\\hline
            TRACE &$0.530$&$0.188$&$0.205$&$0.184$\\\hline
            Transformer & $0.632$ & $0.368$& $0.406$&$0.376$ \\\hline 
            LR &$0.691$&$0.406$&$0.446$&$0.420$\\\hline
            FFNN$_1$ &$0.714$&$0.385$&$0.458$&$0.416$ \\\hline
            RF &$0.738$&$0.391$&$0.504$&$0.434$ \\\hline
            PREVISE &$0.745$&$0.400$&$0.575$&$0.435$\\\hline
            RNN & $0.800$ & $0.494$& $0.583$&$0.511$ \\\hline 
            LSTM & $0.802$ &$0.496$&$0.590$&$0.513$ \\\hline 
        REPLETE &$\mathbf{0.832}$&$\mathbf{0.539}$&$\mathbf{0.615}$&$\mathbf{0.567}$\\\hline
		\end{tabular}
	\label{tbl:sota}
\end{table}

\begin{table}[t]
  \centering
    \caption{Statistical significance comparison between REPLETE and the baselines. Parameter $N$ is set to be $3$.}
	\begin{tabular}{|p{1.5cm}|p{1.2cm}|p{1.2cm}|p{1.2cm}|p{1.2cm}|}
		\hline
           t-score (p-value) & Accuracy & Recall & Precision & $F_1$ score \\\hline 
           w.r.t. PREVISE & $-15.3 $ $ (2.26 \times 10^{-11})$ & $-3.52$ $ (0.0052)$ & $-5.37$ $(8.8 \times 10^{-5})$ & $-4.07 $ $(0.0018)$ \\\hline 
w.r.t. LSTM & $-5.65 $ $ (2.3 \times 10^{-5})$ & $-1.99 $ $(0.064)$ & $-1.86 $ $(0.082)$ & $-2.56$ $ (0.020)$ \\\hline  

		\end{tabular}
	\label{tbl:t-test}
\end{table}

\begin{figure}[t!]
    \centerline{\includegraphics[width=\columnwidth]{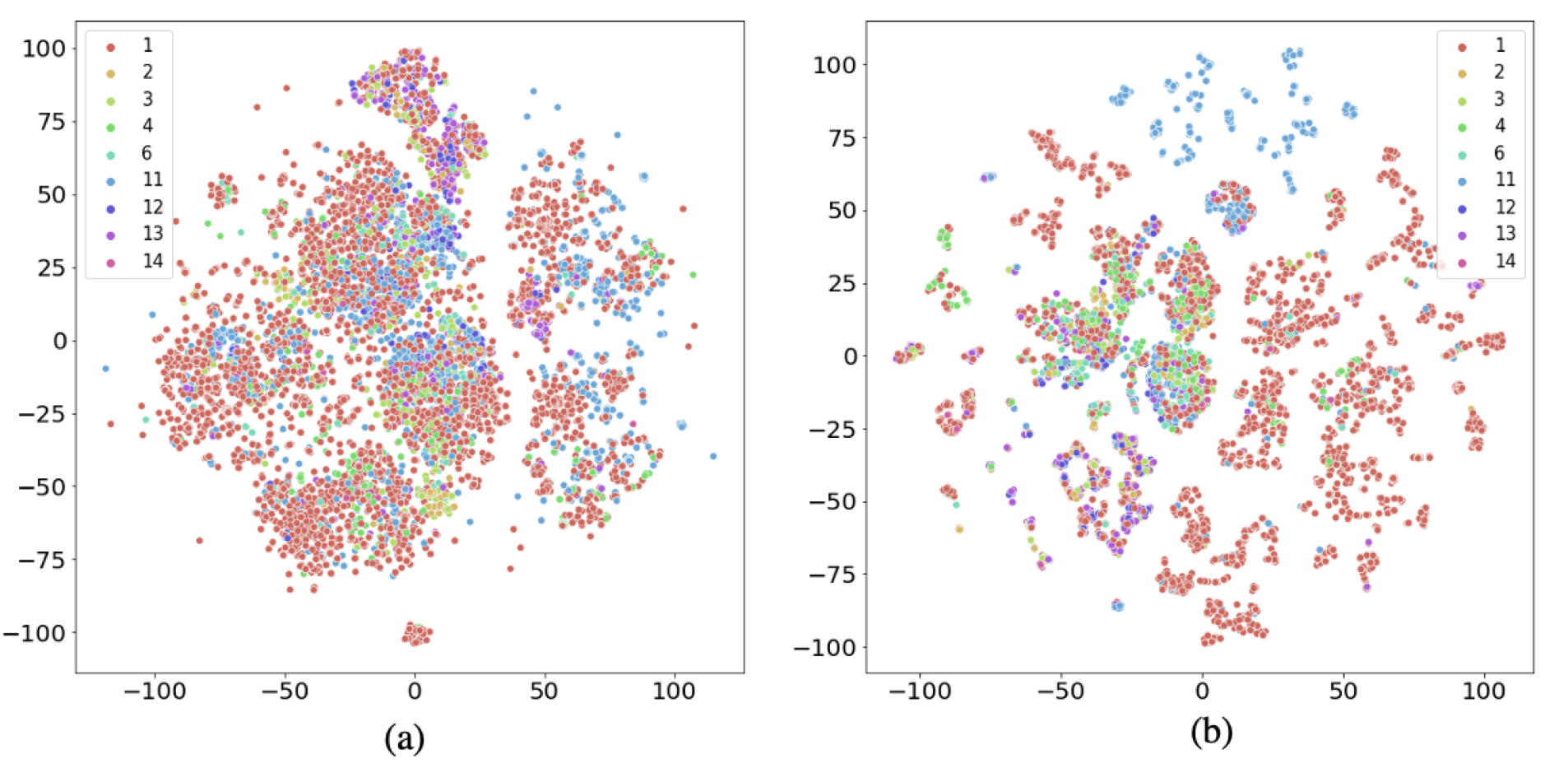}}
    \caption{Scatterplot of 2--dimensional t--SNE embedding for $(a)$ one--hot encoded features  and $(b)$ derived features from learned representations.} 
    \label{fig:tsne}
\end{figure}

To analyze the reliability of the representation learning framework,  we first examine whether derived features can effectively distinguish between labels (i.e., services) compared to the one--hot encoded features. We utilize the t--distributed Stochastic Neighbor Embedding (t--SNE) method to embed both high--dimensional features into 2--dimensional space. Figure~\ref{fig:tsne} illustrates that distinct clusters for each label are formed using the derived features. This indicates that the derived features offer better separability compared to the one--hot encoded features.

Next, we analyze the performance of REPLETE across different history length $N$. Table \ref{tbl:N} shows that our model's performance remain consistent despite variation in the length of the individual's history considered for prediction. This indicates the robustness of our approach. For the rest of our experiments, we set $N = 3$, as the model achieves the highest accuracy and $F_1$ score at this value.
\begin{table}[t]
  \centering
    \caption{Performance of our approach across different history length $N$.}
	\begin{tabular}{|c|c|c|c|c|}
		\hline
		  $N$ & Accuracy & Recall & Precision & $F_1$ score\\\hline
            $2$ & $0.828$& $0.517$ &$\mathbf{0.644}$&$0.547$ \\\hline
            $3$ & $\mathbf{0.832}$&$0.539$&$0.615$&$\mathbf{0.567}$ \\\hline
            $4$ &$0.809$ & $0.499$ & $0.571$ & $0.517$ \\\hline
            $5$ &$0.805$&$0.499$&$0.566$&$0.516$ \\\hline
            $6$ &$0.811$&$\mathbf{0.548}$&$0.607$&$0.561$ \\\hline
		\end{tabular}
	\label{tbl:N}
\end{table}

\paragraph{Ablation Study}
We conducted two ablation studies to evaluate the importance of different model components, focusing on the impact of feature and instance interactions. First, we analyze the impact of feature interactions and instance interactions on model performance. Table \ref{tbl:derive-abl} shows that service and feature representations have the greatest impact, with feature and instance interactions enhancing performance when coupled with these representations.

\begin{table}[t!]
  \centering
    \caption{Ablation study for key components of the derived features, namely service and feature representations (rep), instance interactions (inst), and feature interactions (feat).}
	\begin{tabular}{|c|c|c|c|c|}
	\hline
	   inst & rep & feat & Accuracy & $F_1$ score \\\hline
        \checkmark &  &  & $0.625$&$0.208$\\\hline
         & \checkmark &  & $0.825$&$0.526$\\\hline
         &  & \checkmark & $0.614$&$0.144$\\\hline
         & \checkmark & \checkmark & $0.829$&$0.547$\\\hline
        \checkmark &  & \checkmark & $0.717$&$0.279$\\\hline
        \checkmark & \checkmark &  & $0.829$&$0.543$\\\hline
        \checkmark & \checkmark & \checkmark & $\mathbf{0.832}$&$\mathbf{0.567}$\\\hline
	\end{tabular}
	\label{tbl:derive-abl}
\end{table}

\begin{table}[t!]
  \centering
    \caption{Ablation study for key components of the optimization function, namely temporal (temp), functional (func), and individual (ind).}
	\begin{tabular}{|c|c|c|c|c|}
	\hline
	   ind & temp & func & Accuracy & $F_1$ score \\\hline
        \checkmark &  &  & $0.798$&$0.450$\\\hline
         & \checkmark &  & $0.789$&$0.419$\\\hline
         &  & \checkmark & $0.714$&$0.398$\\\hline
         & \checkmark & \checkmark & $0.801$&$0.476$\\\hline
        \checkmark &  & \checkmark & $0.821$&$0.534$\\\hline
        \checkmark & \checkmark &  & $0.827$&$0.549$\\\hline
        \checkmark & \checkmark & \checkmark & $\mathbf{0.832}$&$\mathbf{0.567}$\\\hline
	\end{tabular}
	\label{tbl:opt-abl}
\end{table}

Next, we analyse the impact of each component of the optimization function on model performance. As shown in Table \ref{tbl:opt-abl}, each component enhances performance, with the best results achieved by combining all three components. 

\paragraph{Hyperparameter Selection}
The joint optimization function has six hyperparameters, with optimal values being $\lambda = 0.01, \alpha = 0.3, \beta = 0.7, \tau = 52$ weeks$, \mu = 1, k =10$. 

\begin{figure}[t!]
    \centerline{\includegraphics[width=\columnwidth]{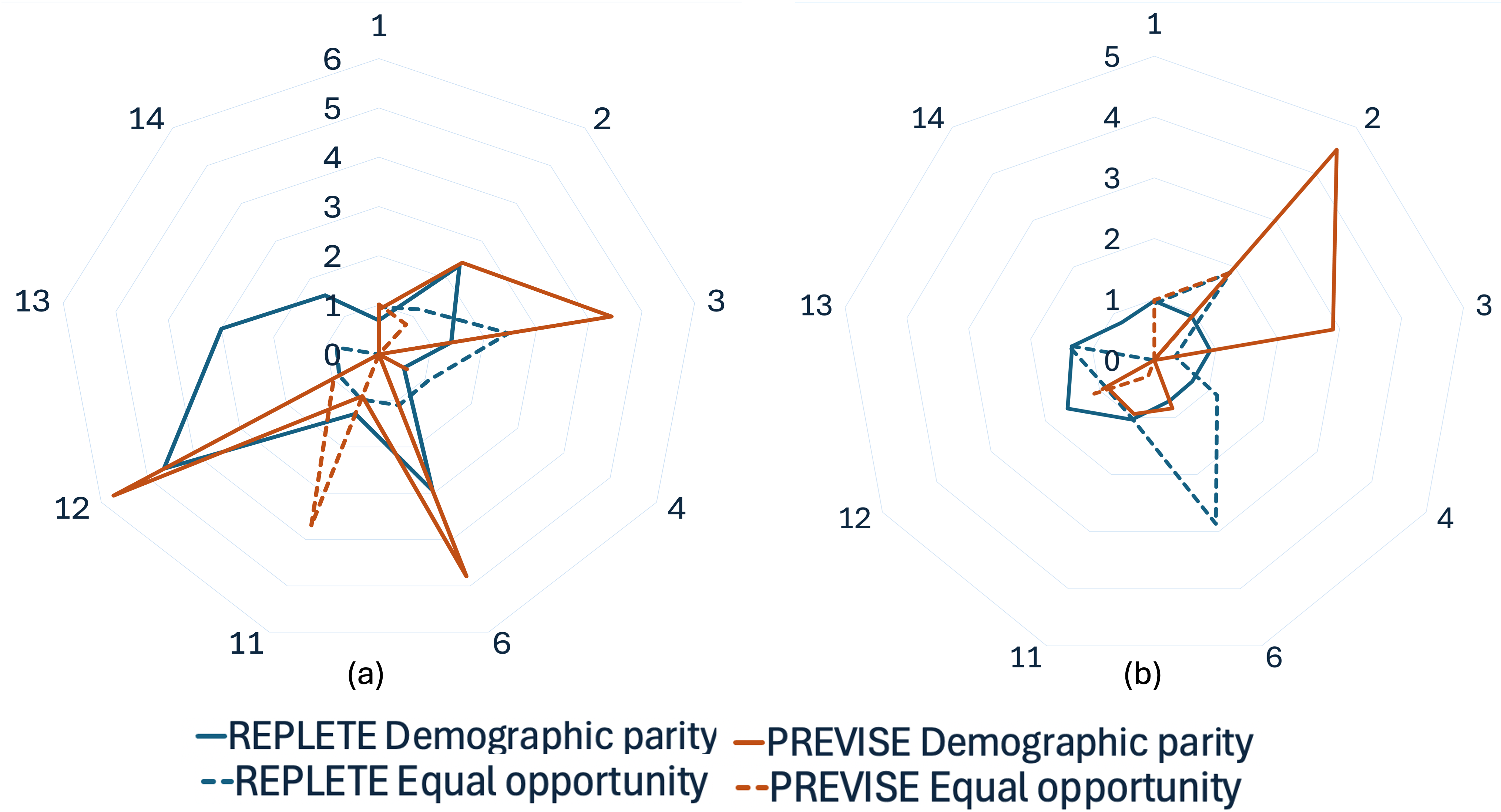}}
    \caption{Demographic parity (solid line) and equal opportunity (dotted line) of sensitive attributes (a) gender and (b) ethnicity for each service using REPLETE (blue) and PREVISE (orange). Plots for the remaining attributes are included in the Appendix section. Numerals are used in lieu of actual service names \cite{2020hmisstandards}.} 
    \label{fig:bias}
\end{figure}

\paragraph{Bias Evaluation}
Although our proposed approach does not explicitly model against bias and unfairness (both critical characteristics of a technological solution to a complex social challenge, such as homelessness service provision), we evaluate it for potential bias across three sensitive attributes, namely race, gender, and ethnicity. We measure bias using $(i)$ demographic parity ($\frac{P(\hat{\mathbf{Y}}=a^i|attr=1)}{P(\hat{\mathbf{Y}}=a^i|attr=0)}$) and $(ii)$ equal opportunity ($\frac{P(\hat{\mathbf{Y}}=a^i|attr=1,\mathbf{Y}=a^i)}{P(\hat{\mathbf{Y}}=a^i|attr=0,\mathbf{Y}=a^i)}$) \cite{mehrabi2021survey}, aiming for values within 
$80\%$ of the group with the highest rate \cite{pessach2022review}. Figure \ref{fig:bias} shows that our approach mitigates bias for high--frequency services, such as emergency shelter (denoted by 1) and day shelter (denoted by 11) for majority of the sensitive attributes, but not as much for lower--frequency services, such as transitional housing (denoted by $2$) and homelessness prevention (denoted by $12$). Moreover, the state--of--the--art method PREVISE exhibits extreme bias for certain services and attributes, where it completely fails to predict some services for specific attributes. This underscores that our approach, REPLETE, demonstrates a significantly lower level of bias compared to PREVISE, ensuring a more equitable service assignment across various attributes and services. In light of this result, we conclude that our approach replicates the existing assignment decision--making process to a significant extent, providing the basis for developing systems that make ``unbiased'' and ``fair'' predictions, as well as to understand and evaluate them ethically (i.e., in experimental settings). We therefore plan to incorporate fairness constraints (e.g., \cite{zafar2019fairness}) directly into our objective function, as part of our future work.

\section{Conclusion}
\label{sec:conclude}
This paper introduced a novel predictive model for homeless service assignment based on representation learning. The proposed approach modeled explicitly both the temporal and functional relationships between services, and the similarity between individuals based on their features and their prior service assignments, to learn latent representations. Utilizing these representations, the proposed approach was shown to outperform the state--of--the--art in the task of next service assignment prediction, a key task in the service assignment decision making process.

\paragraph{Limitations}
Our analysis is based on a geographically bounded dataset, specifically limited to the Capital Region of New York state. Additionally, the dataset does not record the availability or capacity of services.

\paragraph{Future Directions} Our approach serves as a foundational building block for developing more advanced service assignment models. By exploring temporal and functional relationships, we have laid the groundwork for capturing complex interactions in service assignment. However, additional dimensions remain worth investigating, potentially offering valuable insights and new directions for research. Importantly, explicitly addressing bias and unfairness is critical, as our experiments demonstrated, especially if this approach is to be applied in real-world scenarios. Our work takes the first step in this direction by providing a foundation for developing "unbiased" and "fair" predictive models and evaluating them ethically in experimental settings.

REPLETE provides a framework for homelessness service assignment, learning representations that reflect local services and population demographics. While not region-specific, our model has been constrained by the lack of access to homelessness data from other geographical regions due to confidentiality and privacy concerns. Beyond homelessness, we are actively evaluating REPLETE’s applicability to other socially significant domains.

In collaboration with CARES of NY, we are working to test REPLETE in practice. However, empirical evaluation requires addressing ethical considerations, as recommendations could have unintended consequences for individuals. Such a study would also require additional IRB approval. Key technical challenges include mitigating biases in both training data and model outputs and integrating capacity and availability constraints into the predictions. To address these, we plan to incorporate fairness constraints into the optimization function and embed capacity and availability considerations into the prediction model. Additionally, to safeguard privacy and confidentiality, CARES of NY will act as the steward for the model and data.

These efforts position REPLETE as a promising tool for ethical and effective service assignment, with potential for broader applications in other socially impactful domains. By addressing these challenges, we aim to advance the development of fair and unbiased predictive models that can contribute to more equitable outcomes in critical decision-making processes.

\section*{Acknowledgment}
 This material is based upon work supported by the National Science Foundation under Grant No. ECCS-1737443.
\section*{Appendix}
\appendix
\section{Proofs of Update Rules}
In this section, we derive the update rules corresponding to equation \ref{eq:admm} in Section \ref{sec:opt-sol} of the paper. Since $L_{2-1}$ norm is not differentiable, directly applying gradient-based methods for optimization is not feasible. To address this, we leverage Lemma \ref{lemma:2,1} in Section \ref{sec:opt-sol}, which provides a closed-form solution for the updates of the matrices $\mathbf{P}$ and $\mathbf{Q}$. 

\paragraph{Update $\mathbf{P}$}
For updating $\mathbf{P}$, we fix the other variables and remove the terms that are irrelevant to $\mathbf{P}$ in \ref{eq:admm} of Section \ref{sec:opt-sol}. We get $\mathcal{O}_\mathbf{P}=\min_{\mathbf{P}} \|\mathbf{P}\|_{2,1}+\langle \mathcal{K},\mathbf{H}-\mathbf{A}\mathbf{C}^T-\mathbf{P}\rangle +\frac{\mu}{2} \|\mathbf{H}-\mathbf{A}\mathbf{C}^T-\mathbf{P}\|_F^2$, where $\langle \cdot,\cdot \rangle$ denotes the dot product. Mathematically, the solution to this function is as follows:
\begin{equation*}
    \begin{split}
        &\min_{\mathbf{P}} \|\mathbf{P}\|_{2,1}+\langle \mathcal{K},\mathbf{H}-\mathbf{A}\mathbf{C}^T-\mathbf{P}\rangle+\frac{\mu}{2} \|\mathbf{H}-\mathbf{A}\mathbf{C}^T-\mathbf{P}\|_F^2\\
    \end{split}
\end{equation*}
Using the dot product $\langle \mathbf{M},\mathbf{M}' \rangle = \sum_{ij} \mathbf{M}_{ij}\mathbf{M}'_{ij}$ for matrices $\mathbf{M}$ and $\mathbf{M}'$ and the Frobenius 
 norm $\|\mathbf{M}\|_F^2 = \sum_{ij}\mathbf{M}_{ij}$, where $\|\mathbf{M}\|_F^2$ denotes the Frobenius norm of matrix $\mathbf{M}$, we derive the following expression.
\begin{equation*}
    \begin{split}
        &\min_{\mathbf{P}}\sum_{i,j} (\mathcal{K}_{ij}(\mathbf{H}-\mathbf{A}\mathbf{C}^T-\mathbf{P})_{ij}+\frac{\mu}{2} (\mathbf{H}-\mathbf{A}\mathbf{C}^T-\mathbf{P})_{ij}^2)\\
        &+\|\mathbf{P}\|_{2,1}\\
    \end{split}
\end{equation*}
Next, we combine the first two terms into perfect square by adding and subtracting $\frac{1}{\mu}\mathcal{K}_{ij}$. After this adjustment, the term $-\frac{1}{\mu}\mathcal{K}_{ij}$ becomes irrelevant to the optimization of $\mathbf{P}$ and is therefore eliminated from the equation.
\begin{equation*}
    \begin{split}
        &\min_{\mathbf{P}}\sum_{i,j} \frac{\mu}{2}((\mathbf{H}-\mathbf{A}\mathbf{C}^T-\mathbf{P})_{ij}^2+\frac{2}{\mu}\mathcal{K}_{ij}(\mathbf{H}-\mathbf{A}\mathbf{C}^T-\mathbf{P})_{ij}\\
        &+ \frac{1}{\mu}\mathcal{K}_{ij}^2-\frac{1}{\mu}\mathcal{K}_{ij}^2)+\|\mathbf{P}\|_{2,1}\\
        =&\min_{\mathbf{P}}\sum_{i,j} \frac{\mu}{2}((\mathbf{H}-\mathbf{A}\mathbf{C}^T-\mathbf{P})_{ij}^2+\frac{2}{\mu}\mathcal{K}_{ij}(\mathbf{H}-\mathbf{A}\mathbf{C}^T-\mathbf{P})_{ij}\\
        &+\frac{1}{\mu}\mathcal{K}_{ij}^2)+\|\mathbf{P}\|_{2,1}\\
        =&\min_{\mathbf{P}}\sum_{i,j} \frac{\mu}{2}((\mathbf{H}-\mathbf{A}\mathbf{C}^T-\mathbf{P})_{ij}+\frac{1}{\mu}\mathcal{K}_{ij})^2+\|\mathbf{P}\|_{2,1}\\
    \end{split}
\end{equation*}
Next, the squared term can be expressed as Frobenius norm.
\begin{equation*}
    \begin{split}
        =&\min_{\mathbf{P}} \frac{\mu}{2}\|\mathbf{H}-\mathbf{A}\mathbf{C}^T-\mathbf{P}+\frac{1}{\mu}\mathcal{K}\|_F^2+\|\mathbf{P}\|_{2,1}\\
        =&\min_{\mathbf{P}} \mu[\frac{1}{2}\|\mathbf{P}-(\mathbf{H}-\mathbf{A}\mathbf{C}^T+\frac{1}{\mu}\mathcal{K})\|_F^2+\frac{1}{\mu}\|\mathbf{P}\|_{2,1}]\\
        =&\min_{\mathbf{P}} \frac{1}{2}\|\mathbf{P}-(\mathbf{H}-\mathbf{A}\mathbf{C}^T+\frac{1}{\mu}\mathcal{K})\|_F^2+\frac{1}{\mu}\|\mathbf{P}\|_{2,1}\\   
    \end{split}
\end{equation*}
Next, applying Lemma~\ref{lemma:2,1}, the optimal solution $\mathbf{P}^*$ for the above equation is obtained in closed form, where $\mathbf{E}^\mathbf{P}=\mathbf{H}-\mathbf{A}\mathbf{C}^T+\frac{1}{\mu}\mathcal{K}$:
\begin{equation*}
        \mathbf{P}_{i,:}^* = \left\{\begin{array}{lr} (1-\frac{1}{\mu\|\mathbf{E}^\mathbf{P}_{i,:}\|})\mathbf{E}^\mathbf{P}_{i,:}, & \|\mathbf{E}^\mathbf{P}_{i,:}\|>\frac{1}{\mu}\\
        0, & \text{otherwise}\end{array}\right.
\end{equation*}
\paragraph{Update $\mathbf{Q}$}
For updating $\mathbf{Q}$, we fix the other variables and remove the terms that are irrelevant to $\mathbf{Q}$ in Eq.~(\ref{eq:admm}). We get $\min_{\mathbf{Q}} \|\mathbf{Q}\|_{2,1}+\langle \mathcal{L},\mathbf{X}-\mathbf{C}\mathbf{V}^T \mathbf{Q}\rangle+\frac{\mu}{2} \|\mathbf{X}-\mathbf{C}\mathbf{V}^T-\mathbf{Q}\|_F^2$. Similar to $\mathbf{P}$, using Lemma~\ref{lemma:2,1}, the optimal solution $\mathbf{Q}^*$ for the above equation is as follows, where $\mathbf{E}^\mathbf{Q}=\mathbf{X}-\mathbf{C}\mathbf{V}^T+\frac{1}{\mu}\mathcal{L}$.
\begin{equation*}
        \mathbf{Q}_{i,:}^* = \left\{\begin{array}{lr} (1-\frac{1}{\mu\|\mathbf{E}^\mathbf{Q}_{i,:}\|})\mathbf{E}^\mathbf{Q}_{i,:}, & \|\mathbf{E}^\mathbf{Q}_{i,:}\|>\frac{1}{\mu}\\
        0, & \text{otherwise}\end{array}\right.
\end{equation*}

\paragraph{Update $\mathbf{S}$}
For updating $\mathbf{S}$, we fix the other variables and remove the terms that are irrelevant to $\mathbf{S}$, we get $\mathcal{O}_\mathbf{S}=\min \|\mathbf{D}-\mathbf{A}\mathbf{S}^T\|_F^2 + \lambda \|\mathbf{S}\|_F^2 + \alpha \|\mathbf{SB'}\|_F^2 - tr(\mathbf{\psi_SS^\top})$, where $\mathbf{\psi_S}$ is the Lagrangian multiplier for $\mathbf{S}\ge0$ and $tr(\cdot)$ denotes trace operator. The partial derivative of $\mathcal{O}_\mathbf{S}$ is as follows:
\begin{equation*}
\begin{split}
    &\frac{1}{2}\frac{d\mathcal{O}_\mathbf{S}}{d\mathbf{S}}= -(\mathbf{D}-\mathbf{AS}^\top)^\top \mathbf{A}+\alpha(\mathbf{S}\mathbf{B}^{'})\mathbf{B}^{'}+\lambda\mathbf{S}-\psi_\mathbf{S}\\
\end{split}
\end{equation*}
To obtain the optimal solution, we set the partial derivative equal to zero and get
\begin{equation*}
\begin{split}
    &\psi_\mathbf{S} = -\mathbf{D}^{\top}\mathbf{A}+\mathbf{SA}^\top\mathbf{A}+\alpha(\mathbf{S}{(\mathbf{B}^{'})}^2)+\lambda\mathbf{S}
\end{split}
\end{equation*}
Next, we use Karush-KuhnTucker complementary condition, that is, $\psi_\mathbf{S}(i,j)\mathbf{S}_{ij}=0$ \cite{boyd2011distributed}, we get,
\begin{equation*}
\begin{split}
    &( \Tilde{\mathbf{S}}_{ij}-\hat{\mathbf{S}}_{ij})\mathbf{S}_{ij}=0\\
    &\Tilde{\mathbf{S}}_{ij}\mathbf{S}_{ij} =\hat{\mathbf{S}}_{ij}\mathbf{S}_{ij}\\
    &\mathbf{S}_{ij} \leftarrow \mathbf{S}_{ij} \frac{\hat{\mathbf{S}}_{ij}}{\Tilde{\mathbf{S}}_{ij} }\\
\end{split}
\end{equation*}    
Here, $\Tilde{\mathbf{S}}_{ij}$ and $\hat{\mathbf{S}}_{ij}$ consists of the positive and negative terms, respectively, as shown below. 
\begin{equation*}
\begin{split}
    &\hat{\mathbf{S}}_{ij} = \mathbf{D}^{\top}\mathbf{A}+[\alpha(\mathbf{S}{(\mathbf{B}^{'})}^2)]^{-}\\
    &\Tilde{\mathbf{S}}_{ij} = \mathbf{SA}^\top\mathbf{A}+\lambda\mathbf{S}+[\alpha(\mathbf{S}{(\mathbf{B}^{'})}^2)]^{+}\\
\end{split}
\end{equation*}
Here, for any matrix $\mathbf{M}$, $(\mathbf{M})^{+} = \frac{ABS(\mathbf(M)+\mathbf{M}}{\mathbf{M}}$ and $(\mathbf{M})^{-} =\frac{ABS(\mathbf(M)-\mathbf{M}}{\mathbf{M}}$ are the positive and negative part of $\mathbf{M}$, respectively. $ABS(\mathbf{M})$ consists of the absolute value of elements in $\mathbf{M}$ \cite{shu2019beyond}. We follow similar steps to compute $\mathbf{R_p}$, $\mathbf{R_s}$, $\mathbf{V}$, $\mathbf{C}$, and $\mathbf{A}$.

\paragraph{Update $\mathbf{R_p}$ and  $\mathbf{R_s}$}
For updating $\mathbf{R_p}$, we fix the other variables and remove the terms that are irrelevant to $\mathbf{R_p}$, we get $\mathcal{O}_{\mathbf{R_p}} = \min \{ \| \mathbf{T} - \mathbf{A R_p}^{\top}\mathbf{ R_s A}^{\top} \|_F^2\} + \lambda \|\mathbf{R_p}\|_F^2-tr(\psi_{\mathbf{R_p}}\mathbf{R_p}^\top)$. The partial derivative of $\mathcal{O}_\mathbf{R_p}$ is as follows:
\begin{equation*}
    \begin{split}
    &\frac{1}{2}\frac{d\mathcal{O}_\mathbf{R_p}}{d\mathbf{R_p}}= -\mathbf{T}\mathbf{A}\mathbf{A}^\top\mathbf{R_s}+\mathbf{A R_p}^{\top}\mathbf{ R_s A}^{\top}\mathbf{A}\mathbf{A}^\top\mathbf{R_s}\\
    &+\lambda \mathbf{R_p}-\psi_\mathbf{R_p}\\
    &\psi_\mathbf{R_p} = -\mathbf{T}\mathbf{A}\mathbf{A}^\top\mathbf{R_s}+\mathbf{A R_p}^{\top}\mathbf{ R_s A}^{\top}\mathbf{A}\mathbf{A}^\top\mathbf{R_s}+\lambda \mathbf{R_p}
\end{split}
\end{equation*}
Using Karush-KuhnTucker complementary condition \cite{boyd2011distributed}, that is, $\psi_\mathbf{R_p}(i,j)\mathbf{R_p}_{ij}=0$, we get,
\begin{equation*}
\begin{split}
    &\mathbf{R}_{\mathbf{p}_{ij}} \leftarrow \mathbf{R}_{\mathbf{p}_{ij}} \frac{\hat{\mathbf{R}}_{\mathbf{p}_{ij}}}{\Tilde{\mathbf{R}}_{\mathbf{p}_{ij}}}\\
    &\hat{\mathbf{R}}_{\mathbf{p}_{ij}} = \mathbf{T}\mathbf{A}\mathbf{A}^\top\mathbf{R_s}+(\mathbf{A R_p}^{\top}\mathbf{ R_s A}^{\top}\mathbf{A}\mathbf{A}^\top\mathbf{R_s})^{-}\\
    &\Tilde{\mathbf{R}}_{\mathbf{p}_{ij}} = \lambda \mathbf{R_p}+(\mathbf{A R_p}^{\top}\mathbf{ R_s A}^{\top}\mathbf{A}\mathbf{A}^\top\mathbf{R_s})^{+}\\
    \end{split}
\end{equation*}
Similar to $\mathbf{R_p}$, we get the following equations for $\mathbf{R_s}$.
\begin{equation*}
\begin{split}
    &\mathbf{R}_{\mathbf{s}_{ij}} \leftarrow \mathbf{R}_{\mathbf{s}_{ij}} \frac{\hat{\mathbf{R}}_{\mathbf{s}_{ij}}}{\Tilde{\mathbf{R}}_{\mathbf{s}_{ij}}}\\
    &\hat{\mathbf{R}}_{\mathbf{s}_{ij}} = \mathbf{T}\mathbf{A}\mathbf{A}^\top\mathbf{R_p}+(\mathbf{A R_p}^{\top}\mathbf{ R_s A}^{\top}\mathbf{A}\mathbf{A}^\top\mathbf{R_p})^{-}\\
    &\Tilde{\mathbf{R}}_{\mathbf{s}_{ij}} = \lambda \mathbf{R_s}+(\mathbf{A R_p}^{\top}\mathbf{ R_s A}^{\top}\mathbf{A}\mathbf{A}^\top\mathbf{R_p})^{+}\\
\end{split}
\end{equation*}
\paragraph{Update $\mathbf{V}$}
For updating $\mathbf{V}$, we fix the other variables and remove the terms that are irrelevant to $\mathbf{V}$, we get $\mathcal{O}_{\mathbf{V}} = \min \frac{\mu}{2}\|\mathbf{X}-\mathbf{CV}^\top-\mathbf{Q} \|_F^2+ \langle \mathcal{L},\mathbf{X}-\mathbf{CV}^\top-\mathbf{Q} \rangle+\lambda\|\mathbf{V}\|_F^2-tr(\psi_\mathbf{V}\mathbf{V}^\top)$. The partial derivative of $\mathcal{O}_\mathbf{V}$ is as follows:
\begin{equation*}
    \begin{split}
    &\frac{d\mathcal{O}_\mathbf{V}}{d\mathbf{V}}= -\mu(\mathbf{X}-\mathbf{CV}^\top-\mathbf{Q})^\top\mathbf{C}-\mathcal{L}^\top\mathbf{C} +2\lambda\mathbf{V} -\psi_\mathbf{V}\\
    &\psi_\mathbf{V} = -\mu\mathbf{X}^\top\mathbf{C}+ \mu\mathbf{VC}^\top\mathbf{C}+\mu\mathbf{Q}^\top\mathbf{C}-\mathcal{L}^\top\mathbf{C} +2\lambda\mathbf{V}
\end{split}
\end{equation*}
Using Karush-KuhnTucker complementary condition  \cite{boyd2011distributed}, that is, $\psi_\mathbf{V}(i,j)\mathbf{V}_{ij}=0$, we get,
\begin{equation*}
\begin{split}
    &\mathbf{V}_{ij} \leftarrow \mathbf{V}_{ij} \frac{\hat{\mathbf{V}}_{ij}}{\Tilde{\mathbf{V}}_{ij}}\\
    &\hat{\mathbf{V}}_{ij} = \mu\mathbf{X}^\top\mathbf{C} + (\mathcal{L}^\top\mathbf{C})^{+}\\
    &\Tilde{\mathbf{V}}_{ij} = \mu\mathbf{VC}^\top\mathbf{C}+\mu\mathbf{Q}^\top\mathbf{C}+(\mathcal{L}^\top\mathbf{C})^{-} +2\lambda\mathbf{V}\\
\end{split}
\end{equation*}
\paragraph{Update $\mathbf{C}$}
For updating $\mathbf{C}$, we fix the other variables and remove the terms that are irrelevant to $\mathbf{C}$, we get $\mathcal{O}_{\mathbf{C}} = \min  \frac{\mu}{2}\|\mathbf{H}-\mathbf{AC}^\top-\mathbf{P} \|_F^2+\frac{\mu}{2}\|\mathbf{X}-\mathbf{CV}^\top-\mathbf{Q} \|_F^2 + \langle \mathcal{L},\mathbf{X}-\mathbf{CV}^\top-\mathbf{Q} \rangle + \langle \mathcal{K},\mathbf{H}-\mathbf{AC}^\top-\mathbf{P} \rangle + \langle \mathcal{N,\mathbf{C^\top C-I}} \rangle+\beta tr(\mathbf{C}^{\top}\mathbf{\Gamma}\mathbf{C})+\lambda\|\mathbf{C}\|_F^2 -tr(\psi_\mathbf{C}\mathbf{C}^\top)$. The partial derivative of $\mathcal{O}_\mathbf{C}$ is as follows:
\begin{equation*}
    \begin{split}
    &\frac{1}{2}\frac{d\mathcal{O}_\mathbf{C}}{d\mathbf{C}}= -\mu(\mathbf{H}-\mathbf{AC}^\top-\mathbf{P})^\top\mathbf{A}-\mu(\mathbf{X}-\mathbf{CV}^\top-\mathbf{Q})\mathbf{V}\\
    &-\mathcal{L}\mathbf{V}-\mathcal{K}^\top\mathbf{A}+2\mathbf{C}\mathcal{N}+2\beta\mathbf{\Gamma}\mathbf{C}+2\lambda\mathbf{C}-\psi_\mathbf{C}\\
    &\psi_\mathbf{C} = -\mu\mathbf{H}^\top\mathbf{A}+\mu\mathbf{C}\mathbf{A}^\top\mathbf{A}+\mu \mathbf{P}^\top\mathbf{A}-\mu\mathbf{X}\mathbf{V}+\mu\mathbf{C}\mathbf{V}^\top\mathbf{V}\\
    &+\mu\mathbf{Q}\mathbf{V}-\mathcal{L}\mathbf{V}-\mathcal{K}^\top\mathbf{A}+2\mathbf{C}\mathcal{N}+2\beta\mathbf{\Gamma}\mathbf{C}+2\lambda\mathbf{C}\\
\end{split}
\end{equation*}
Using Karush-KuhnTucker complementary condition  \cite{boyd2011distributed}, that is, $\psi_\mathbf{C}(i,j)\mathbf{C}_{ij}=0$, we get,
\begin{equation*}
\begin{split}
    &\mathbf{C}_{ij} \leftarrow \mathbf{C}_{ij} \frac{\hat{\mathbf{C}}_{ij}}{\Tilde{\mathbf{C}_{ij}}}\\
    &\hat{\mathbf{C}}_{ij} = \mu\mathbf{H}^\top\mathbf{A}+\mu\mathbf{X}\mathbf{V}+(\mathcal{L}\mathbf{V})^{+}+(\mathcal{K}^\top\mathbf{A})^{+}+2(\mathbf{C}\mathcal{N})^{-}\\
    &+2(\beta(\mathbf{\Gamma}\mathbf{C})^{-}\\
    &\Tilde{\mathbf{C}}_{ij} = \mu\mathbf{C}\mathbf{A}^\top\mathbf{A}+\mu \mathbf{P}^\top\mathbf{A}+\mu\mathbf{C}\mathbf{V}^\top\mathbf{V}+\mu\mathbf{Q}\mathbf{V}+(\mathcal{L}\mathbf{V})^{-}\\
    &+(\mathcal{K}^\top\mathbf{A})^{-} +2(\mathbf{C}\mathcal{N})^{+}+2\beta(\mathbf{\Gamma}\mathbf{C})^{+}+2\lambda\mathbf{C}\\
\end{split}
\end{equation*}
\paragraph{Update $\mathbf{A}$}
For updating $\mathbf{A}$, we fix the other variables and remove the terms that are irrelevant to $\mathbf{A}$, we get $\mathcal{O}_{\mathbf{A}} = \| \mathbf{D} - \mathbf{AS}^{\top} \|_F^2 +\| \mathbf{T} - \mathbf{A R_p}^{\top}\mathbf{ R_s A}^{\top} \|_F^2+\lambda \|\mathbf{A}\|_F^2+ \frac{\mu}{2}\|\mathbf{H}-\mathbf{AC}^\top-\mathbf{P} \|_F^2+ \langle \mathcal{K},\mathbf{H}-\mathbf{AC}^\top-\mathbf{P} \rangle-tr(\psi_\mathbf{A}\mathbf{A}^\top) \min $. The partial derivative of $\mathcal{O}_\mathbf{A}$ is as follows:
\begin{equation*}
    \begin{split}
    &\frac{1}{2}\frac{d\mathcal{O}_\mathbf{A}}{d\mathbf{A}}= -2(\mathbf{D}-\mathbf{AS}^\top)\mathbf{S}-4(\mathbf{T}-\mathbf{A R_p}^{\top}\mathbf{ R_s A}^{\top} )\mathbf{A R_p}^{\top}\mathbf{ R_s } \\
    &+2\lambda\mathbf{A}-\mu(\mathbf{H}-\mathbf{A}\mathbf{C}^\top-\mathbf{P})\mathbf{C}-\mathbf{K}\mathbf{C}-\psi_\mathbf{A}\\
    &\psi_\mathbf{A} = - 2\mathbf{D}\mathbf{S}+2 \mathbf{AS}^\top\mathbf{S}-4\textbf{T}\mathbf{A}\mathbf{R_p}^\top\mathbf{R_s}\\
    &+4\mathbf{A R_p}^{\top}\mathbf{ R_s A}^{\top}\mathbf{A R_p}^{\top}\mathbf{ R_s }+2\lambda\mathbf{A}-\mu\mathbf{H}\mathbf{C}+\mu\mathbf{A}\mathbf{C}^\top\mathbf{C}\\
    &+\mu\mathbf{P}\mathbf{C}-\mathbf{K}\mathbf{C}
\end{split}
\end{equation*}
Using Karush-KuhnTucker complementary condition  \cite{boyd2011distributed}, that is, $\psi_\mathbf{A}(i,j)\mathbf{A}_{ij}=0$, we get,
\begin{equation*}
\begin{split}
    &\mathbf{A}_{ij} \leftarrow \mathbf{A}_{ij} \frac{\hat{\mathbf{A}}_{ij}}{\Tilde{\mathbf{A}_{ij}}}\\
    &\hat{\mathbf{A}}_{ij} =2\mathbf{D}\mathbf{S}+4\textbf{T}\mathbf{A}\mathbf{R_p}^\top\mathbf{R_s}+4(\mathbf{A R_p}^{\top}\mathbf{ R_s A}^{\top}\mathbf{A R_p}^{\top}\mathbf{ R_s })^{-}\\ &+\mu\mathbf{H}\mathbf{C}+ (\mathbf{K}\mathbf{C})^{+}\\
    &\Tilde{\mathbf{A}}_{ij} = 2 \mathbf{AS}^\top\mathbf{S}+4(\mathbf{A R_p}^{\top}\mathbf{ R_s A}^{\top}\mathbf{A R_p}^{\top}\mathbf{ R_s })^{+}+2\lambda\mathbf{A}\\   &+\mu\mathbf{A}\mathbf{C}^\top\mathbf{C}+\mu\mathbf{P}\mathbf{C}+(\mathbf{K}\mathbf{C})^{-}\\
\end{split}
\end{equation*}

\section{Further Experimental Results on Bias}
\begin{figure}[H] \centerline{\includegraphics[width=\columnwidth]{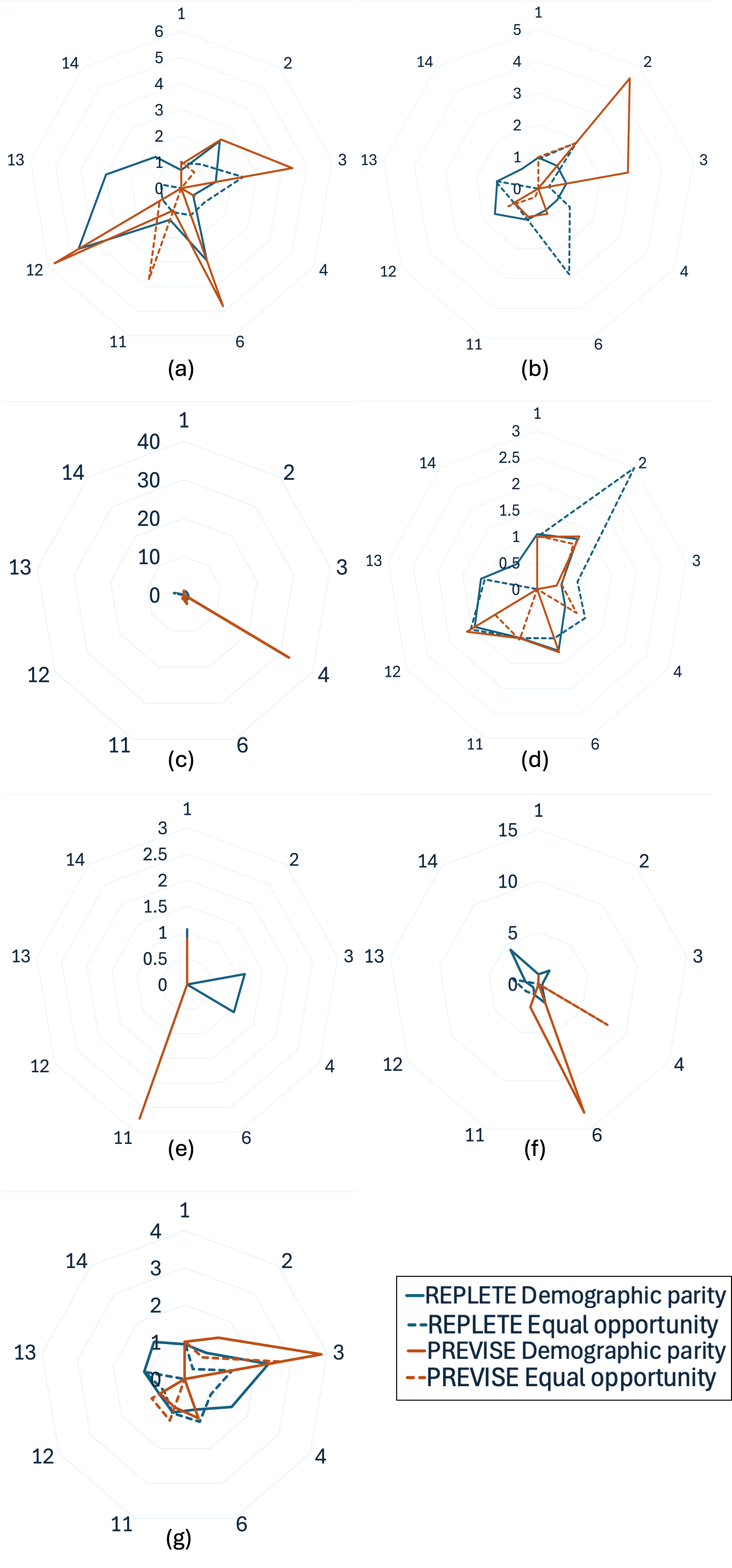}}
    \caption{Demographic parity (solid line) and equal opportunity (dotted line) of sensitive attributes (a) gender, (b) ethnicity, (c) American Indian or Alaskan Native, (d) Black African American, (e) Asian, (f) Native Hawaiian or Other Pacific Islander, and (g) White  for each service using REPLETE (blue) and PREVISE (orange). Numerals are used in lieu of actual service names \cite{2020hmisstandards}. Values within the range of $0.8$ to $1.2$ are consider better, with the optimal value being $1$ \cite{pessach2022review}.} 
    \label{fig:bias-extra}
\end{figure}
In Section \ref{sec:results}, Figure~\ref{fig:bias} compares our proposed approach (REPLETE) with the state--of--the--art (PREVISE) w.r.t demographic parity and equal opportunity for two sensitive attributes, namely gender and ethnicity. Here, we plot additional results for more sensitive attributes in our dataset. We observe that PREVISE fails to predict certain services for specific attributes altogether, resulting in either zero or disproportionately high values (closer to 15 or 32) for demographic parity and equal opportunity. A value of zero indicates that no individuals with that specific attribute were assigned to the service, suggesting bias against the minority group. Conversely, higher values suggest that the model is biased against the majority group. In contrast, REPLETE exhibits a more balanced behavior, avoiding these extremes and making relatively unbiased assignments.

\section{Further Performance Comparison}
To evaluate the significance of the features derived from the representation learning framework of REPLETE (Section~\ref{sec:learningframework}), we compare our model against an additional baseline, where sequences of one-hot encoded features are input into the feed-forward neural network (FFNN$_1$) instead of the derived features. Table \ref{tbl:sota-add} shows that naively training a FFNN on the one-hot encoded features results in a model with a predictive power that is better than TRACE and LR but not as good as RF, and significantly worse than both PREVISE and REPLETE. In fact REPLETE (which uses FFNN with the learned representations) achieves the best predictive performance, demonstrating that the derived features are indeed crucial in accurately predicting the next service assignment.
\begin{table}[H]
  \centering
    \caption{Performance comparison between REPLETE and FFNN$_1$. Parameter $N$ is set to be $3$.}
	\begin{tabular}{|p{1.5cm}|c|c|c|c|}
		\hline
		  Method & Accuracy & Recall & Precision & $F_1$ score\\\hline
            FFNN$_1$ &$0.714$&$0.385$&$0.458$&$0.416$ \\\hline
            REPLETE &$\mathbf{0.832}$&$\mathbf{0.539}$&$\mathbf{0.615}$&$\mathbf{0.567}$\\\hline
		\end{tabular}
	\label{tbl:sota-add}
\end{table}

\section{Clarification on Features used to Train the Baselines}
\begin{figure}[H]
    \centerline{\includegraphics[width=\columnwidth]{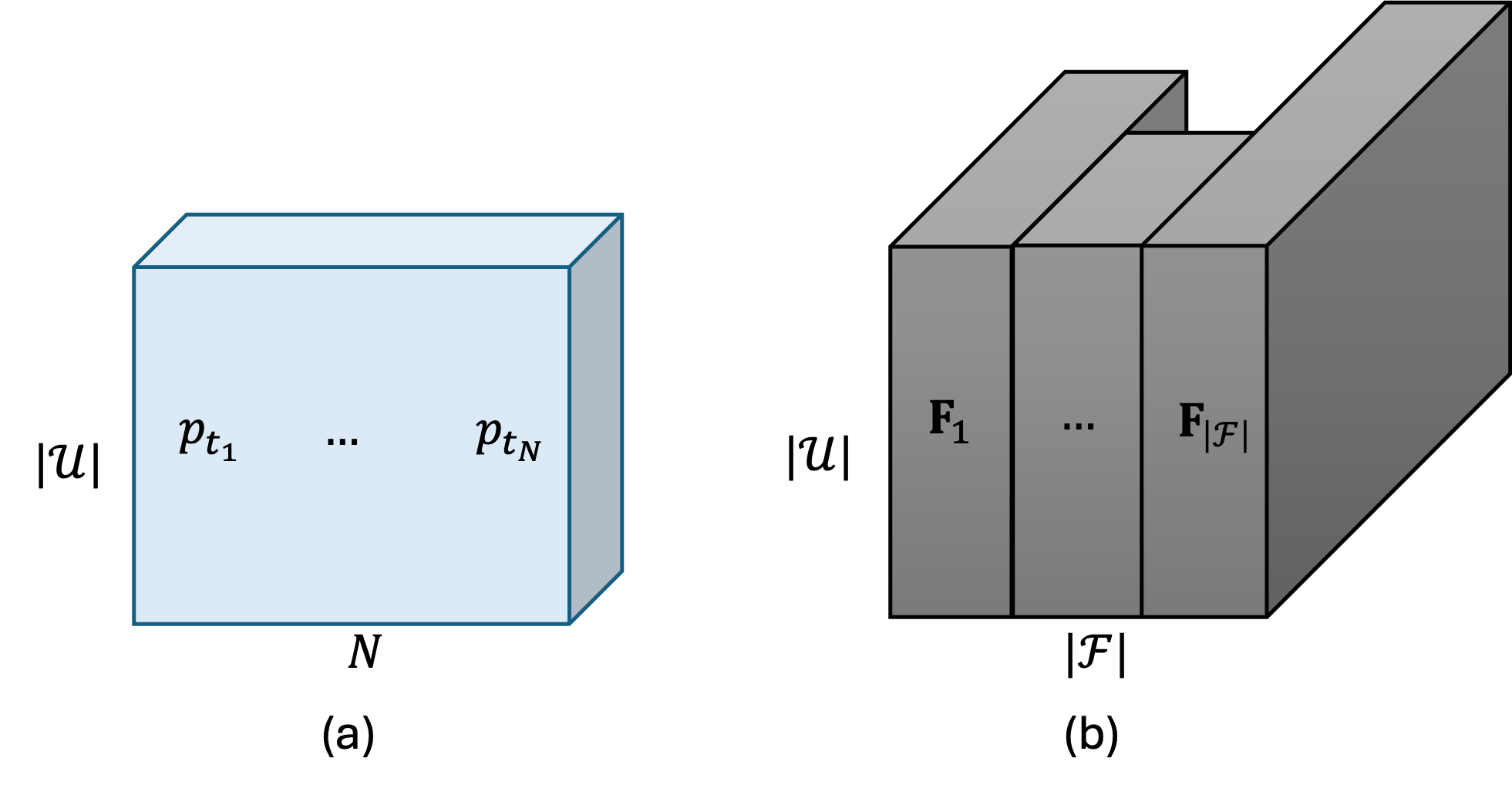}}
    \caption{Input for (a) TRACE and PREVISE, and (b) RF, LR, FFNN$_1$. As defined in Section \ref{sec:problem}, $\mathcal{U}$, $\mathcal{F}$, and $p_{t_i}$ is the set of chronically homeless individuals, set of features, and service assigned at time $t_i$, respectively. Additionally, $\mathbf{F}_i$ denotes the one--hot encoded vector of feature $\mathcal{F}_i$.} 
    \label{fig:input}
\end{figure}

Figure \ref{fig:input} provides a visual representation of the inputs used by the baselines. PREVISE and TRACE directly use the sequence of services $T_{u_i}$ for each individual $u_i$ and leverage the transition between services \cite{rahman2022learning} within a network (Bayesian network \cite{rahman2023bayesian}) for service prediction. On the other hand, RF, LR, and FFNN$_1$ all use a concatenated vector of one-hot encoded features (e.g., gender, living situation, disability condition etc.) as their input.

\section{Hyperparameter Tuning}
\begin{figure}[t!]
    \centerline{\includegraphics[width=\columnwidth]{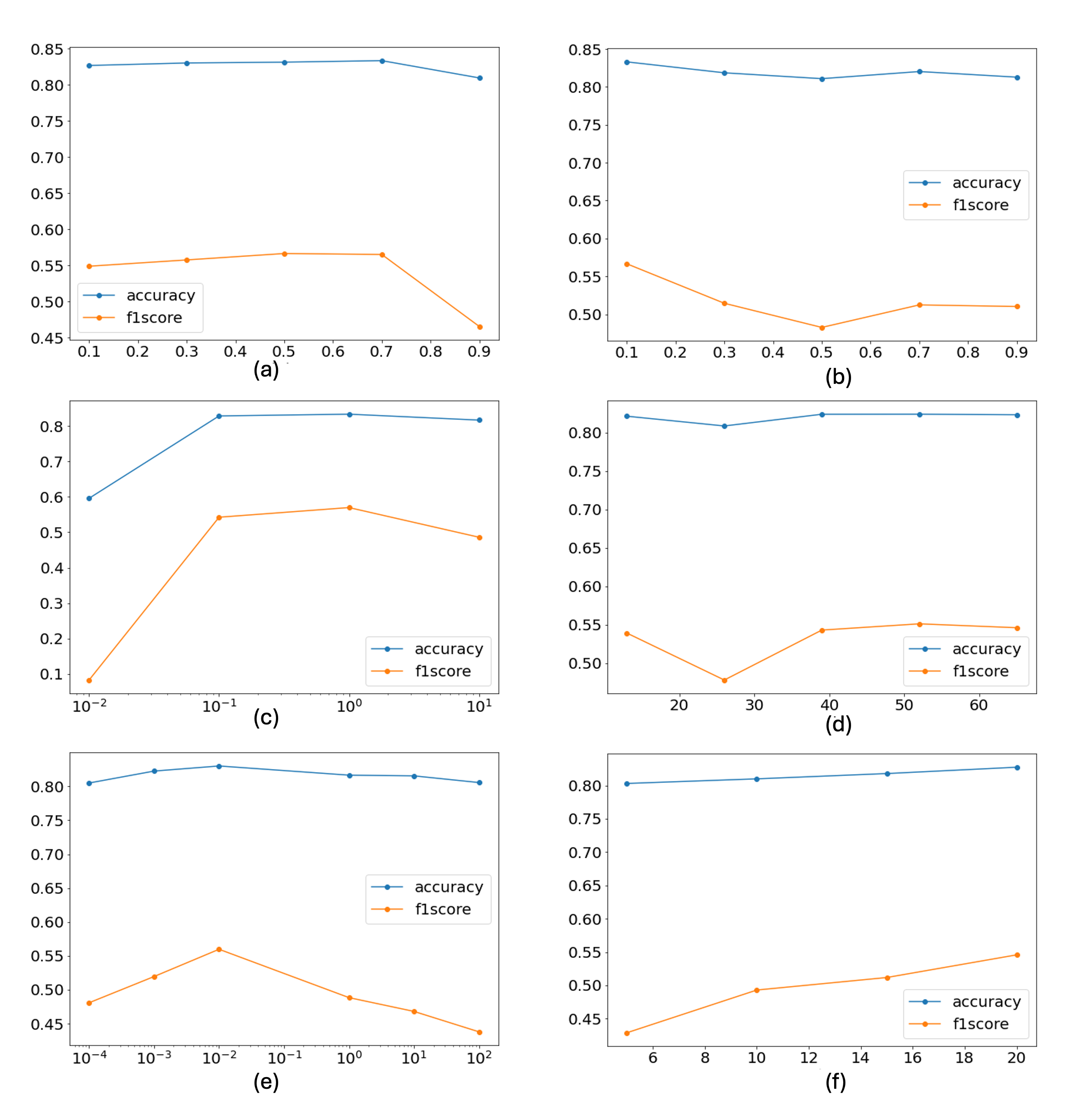}}
    \caption{Accuracy and $F_1$ score plots for varying $(a)$ $\alpha$, $(b)$ $\beta$, $(c)$ $\mu$, $(d)$ $\tau$, $(e)$ $\lambda$, and $(f)$ $k$ .} 
    \label{fig:hyperparameter}
\end{figure}

We assessed their impact on model performance, finding that our approach consistently performs well across different hyperparameter values illustrated in Figure \ref{fig:hyperparameter}.

\bibstyle{aaai}
\bibliography{biblio}

\begin{thebibliography}{29}
\providecommand{\natexlab}[1]{#1}

\bibitem[{Boyd et~al.(2011)Boyd, Parikh, Chu, Peleato, Eckstein et~al.}]{boyd2011distributed}
Boyd, S.; Parikh, N.; Chu, E.; Peleato, B.; Eckstein, J.; et~al. 2011.
\newblock Distributed optimization and statistical learning via the alternating direction method of multipliers.
\newblock \emph{Foundations and Trends{\textregistered} in Machine learning}, 3(1): 1--122.

\bibitem[{Boyd and Vandenberghe(2004)}]{boyd2004convex}
Boyd, S.; and Vandenberghe, L. 2004.
\newblock \emph{Convex optimization}.
\newblock Cambridge university press.

\bibitem[{Chelmis et~al.(2021)Chelmis, Qi, Lee, and Duncan}]{chelmis2021smart}
Chelmis, C.; Qi, W.; Lee, W.; and Duncan, S. 2021.
\newblock Smart Homelessness Service Provision with Machine Learning.
\newblock \emph{Procedia Computer Science}, 185: 9--18.

\bibitem[{Dej, Gaetz, and Schwan(2020)}]{dej2020turning}
Dej, E.; Gaetz, S.; and Schwan, K. 2020.
\newblock Turning off the tap: a typology for homelessness prevention.
\newblock \emph{The Journal of Primary Prevention}, 41(5): 397--412.

\bibitem[{Fleury et~al.(2021)Fleury, Grenier, Sabetti, Bertrand, Cl{\'e}ment, and Brochu}]{fleury2021met}
Fleury, M.-J.; Grenier, G.; Sabetti, J.; Bertrand, K.; Cl{\'e}ment, M.; and Brochu, S. 2021.
\newblock Met and unmet needs of homeless individuals at different stages of housing reintegration: A mixed-method investigation.
\newblock \emph{PloS one}, 16(1): e0245088.

\bibitem[{Gao, Das, and Fowler(2017)}]{gao2017homelessness}
Gao, Y.; Das, S.; and Fowler, P. 2017.
\newblock Homelessness service provision: a data science perspective.
\newblock In \emph{Workshops at the Thirty-First AAAI Conference on Artificial Intelligence}.

\bibitem[{Greer et~al.(2016)Greer, Shinn, Kwon, and Zuiderveen}]{greer2016targeting}
Greer, A.~L.; Shinn, M.; Kwon, J.; and Zuiderveen, S. 2016.
\newblock Targeting services to individuals most likely to enter shelter: Evaluating the efficiency of homelessness prevention.
\newblock \emph{Social Service Review}, 90(1): 130--155.

\bibitem[{Henry et~al.(2023)Henry, de~Sousa, Roddey, Gayen, Joe~Bednar, and Associates}]{ahar2023}
Henry, M.; de~Sousa, T.; Roddey, C.; Gayen, S.; Joe~Bednar, T.; and Associates, A. 2023.
\newblock {The 2023 Annual Homeless Assessment Report (AHAR) to Congress. Part 1: Point-in-time estimates of homelessness}.
\newblock \emph{{The US Department of Housing and Urban Development}}.

\bibitem[{Hong et~al.(2018)Hong, Malik, Lundquist, Bellach, and Kontokosta}]{hong2018applications}
Hong, B.; Malik, A.; Lundquist, J.; Bellach, I.; and Kontokosta, C.~E. 2018.
\newblock Applications of machine learning methods to predict readmission and length-of-stay for homeless families: The case of win shelters in new york city.
\newblock \emph{Journal of Technology in Human Services}, 36(1): 89--104.

\bibitem[{Kuang, Ding, and Park(2012)}]{kuang2012symmetric}
Kuang, D.; Ding, C.; and Park, H. 2012.
\newblock Symmetric nonnegative matrix factorization for graph clustering.
\newblock In \emph{Proceedings of the 2012 SIAM international conference on data mining}, 106--117. SIAM.

\bibitem[{Kube, Das, and Fowler(2019)}]{kube2019allocating}
Kube, A.; Das, S.; and Fowler, P.~J. 2019.
\newblock Allocating interventions based on predicted outcomes: A case study on homelessness services.
\newblock In \emph{Proceedings of the AAAI Conference on Artificial Intelligence}, volume~33, 622--629.

\bibitem[{Mehrabi et~al.(2021)Mehrabi, Morstatter, Saxena, Lerman, and Galstyan}]{mehrabi2021survey}
Mehrabi, N.; Morstatter, F.; Saxena, N.; Lerman, K.; and Galstyan, A. 2021.
\newblock A survey on bias and fairness in machine learning.
\newblock \emph{ACM computing surveys (CSUR)}, 54(6): 1--35.

\bibitem[{Messier, John, and Malik(2021)}]{messier2021predicting}
Messier, G.; John, C.; and Malik, A. 2021.
\newblock Predicting Chronic Homelessness: The Importance of Comparing Algorithms using Client Histories.
\newblock \emph{Journal of Technology in Human Services}, 1--12.

\bibitem[{Pessach and Shmueli(2022)}]{pessach2022review}
Pessach, D.; and Shmueli, E. 2022.
\newblock A review on fairness in machine learning.
\newblock \emph{ACM Computing Surveys (CSUR)}, 55(3): 1--44.

\bibitem[{Pokharel, Das, and Fowler(2024)}]{pokharel2024discretionary}
Pokharel, G.; Das, S.; and Fowler, P. 2024.
\newblock Discretionary trees: understanding street-level bureaucracy via machine learning.
\newblock In \emph{Proceedings of the AAAI Conference on Artificial Intelligence}, volume~38, 22303--22312.

\bibitem[{Rahman and Chelmis(2022)}]{rahman2022learning}
Rahman, K.~S.; and Chelmis, C. 2022.
\newblock Learning to Predict Transitions within the Homelessness System from Network Trajectories.
\newblock In \emph{Proceedings of the 2022 IEEE/ACM International Conference on Advances in Social Networks Analysis and Mining}, 181--189.

\bibitem[{Rahman, Zois, and Chelmis(2023)}]{rahman2023bayesian}
Rahman, K.~S.; Zois, D.-S.; and Chelmis, C. 2023.
\newblock Bayesian Network Modeling and Prediction of Transitions Within the Homelessness System.
\newblock In \emph{ICASSP 2023-2023 IEEE International Conference on Acoustics, Speech and Signal Processing (ICASSP)}, 1--5. IEEE.

\bibitem[{Rodr{\'\i}guez et~al.(2018)Rodr{\'\i}guez, Bautista, Gonzalez, and Escalera}]{rodriguez2018beyond}
Rodr{\'\i}guez, P.; Bautista, M.~A.; Gonzalez, J.; and Escalera, S. 2018.
\newblock Beyond one-hot encoding: Lower dimensional target embedding.
\newblock \emph{Image and Vision Computing}, 75: 21--31.

\bibitem[{Sarker(2021)}]{sarker2021machine}
Sarker, I.~H. 2021.
\newblock Machine learning: Algorithms, real-world applications and research directions.
\newblock \emph{SN computer science}, 2(3): 160.

\bibitem[{Shinn et~al.(2013)Shinn, Greer, Bainbridge, Kwon, and Zuiderveen}]{shinn2013efficient}
Shinn, M.; Greer, A.~L.; Bainbridge, J.; Kwon, J.; and Zuiderveen, S. 2013.
\newblock Efficient targeting of homelessness prevention services for families.
\newblock \emph{American journal of public health}, 103(S2): S324--S330.

\bibitem[{Shu, Wang, and Liu(2019)}]{shu2019beyond}
Shu, K.; Wang, S.; and Liu, H. 2019.
\newblock Beyond news contents: The role of social context for fake news detection.
\newblock In \emph{Proceedings of the twelfth ACM international conference on web search and data mining}, 312--320.

\bibitem[{Tang and Liu(2012)}]{tang2012unsupervised}
Tang, J.; and Liu, H. 2012.
\newblock Unsupervised feature selection for linked social media data.
\newblock In \emph{Proceedings of the 18th ACM SIGKDD international conference on Knowledge discovery and data mining}, 904--912.

\bibitem[{Toros and Flaming(2018)}]{toros2018prioritizing}
Toros, H.; and Flaming, D. 2018.
\newblock Prioritizing homeless assistance using predictive algorithms: an evidence-based approach.
\newblock \emph{Cityscape}, 20(1): 117--146.

\bibitem[{{United States Department of Housing and Urban Development}(2020)}]{2020hmisstandards}
{United States Department of Housing and Urban Development}. 2020.
\newblock {HMIS Data Standards Manual}.
\newblock \emph{Retrieved May 26, 2021, from \url{https://www.hudexchange.info/resource/3824/hmis-data-dictionary/}}.

\bibitem[{Vajiac et~al.(2024)Vajiac, Frey, Baumann, Smith, Amarasinghe, Lai, Rodolfa, and Ghani}]{vajiac2024preventing}
Vajiac, C.; Frey, A.; Baumann, J.; Smith, A.; Amarasinghe, K.; Lai, A.; Rodolfa, K.~T.; and Ghani, R. 2024.
\newblock Preventing Eviction-Caused Homelessness through ML-Informed Distribution of Rental Assistance.
\newblock In \emph{Proceedings of the AAAI Conference on Artificial Intelligence}, volume~38, 22393--22400.

\bibitem[{VanBerlo et~al.(2021)VanBerlo, Ross, Rivard, and Booker}]{vanberlo2021interpretable}
VanBerlo, B.; Ross, M.~A.; Rivard, J.; and Booker, R. 2021.
\newblock Interpretable machine learning approaches to prediction of chronic homelessness.
\newblock \emph{Engineering Applications of Artificial Intelligence}, 102: 104243.

\bibitem[{Wang, Tang, and Liu(2015)}]{wang2015embedded}
Wang, S.; Tang, J.; and Liu, H. 2015.
\newblock Embedded unsupervised feature selection.
\newblock In \emph{Proceedings of the AAAI conference on artificial intelligence}, volume~29.

\bibitem[{Zafar et~al.(2019)Zafar, Valera, Gomez-Rodriguez, and Gummadi}]{zafar2019fairness}
Zafar, M.~B.; Valera, I.; Gomez-Rodriguez, M.; and Gummadi, K.~P. 2019.
\newblock Fairness constraints: A flexible approach for fair classification.
\newblock \emph{Journal of Machine Learning Research}, 20(75): 1--42.

\bibitem[{Zhang and Zhang(2017)}]{zhang2017unified}
Zhang, L.; and Zhang, S. 2017.
\newblock A unified joint matrix factorization framework for data integration.
\newblock \emph{arXiv preprint arXiv:1707.08183}.

\end{thebibliography}

\end{document}